\documentclass[lettersize,journal]{IEEEtran}
\usepackage{amsmath,amsfonts}
\usepackage{algorithmic}
\usepackage{algorithm}
\usepackage{array}
\usepackage[caption=false,font=normalsize,labelfont=sf,textfont=sf]{subfig}
\usepackage{textcomp}
\usepackage{stfloats}
\usepackage{url}
\usepackage{verbatim}
\usepackage{graphicx}
\usepackage{cite}
\usepackage{svg}
\usepackage{multirow}
\usepackage{bbding}
\usepackage{xcolor}
\usepackage{adjustbox}
\usepackage{orcidlink}

\begin{document}

\title{A Stepwise Distillation Learning Strategy for Non-differentiable Visual Programming Frameworks on Visual Reasoning Tasks}

\author{Wentao~Wan~\orcidlink{0009-0004-2063-4382}$^{*}$, 
        Nan~Kang$^{*}$, 
        Zhuojie~Yang, 
        Zeqing~Wang~\orcidlink{0009-0006-6389-6678}, 
        Keze~Wang~\orcidlink{0000-0002-7817-8306}$^{\dagger}$, 
        Liang~Lin~\orcidlink{0000-0003-2248-3755}, \IEEEmembership{Fellow, IEEE}

\thanks{$^{*}$These authors contributed equally to this work.}
\thanks{$^{\dagger}$Keze Wang is the corresponding author.}
\thanks{Wentao Wan, Nan Kang, Zhuojie Yang, Zeqing Wang, Liang Lin, and Keze Wang are with the School of Computer Science and Engineering, Sun Yat-sen University, Guangzhou, Guangdong 510006, China (E-mail: wanwt3@mail2.sysu.edu.cn; kezewang@gmail.com; lianglin@ieee.org).}
}

\markboth{Submitted to IEEE Transactions on Image Processing}
{Shell \MakeLowercase{\textit{et al.}}: A Sample Article Using IEEEtran.cls for IEEE Journals}


\maketitle

\begin{abstract}
Recently, Visual Programming (VProg) has emerged as a significant framework for visual reasoning (VR) tasks due to its interpretability and cross-task generality. However, even with invoking powerful pre-trained Vision-Language models (VLMs) as visual sub-modules, the performance of VProg on specific VR tasks is markedly inferior compared to well-trained task-specific networks.
Although invoking task-specific models can further enhance the performance of VProg on specific VR tasks, it greatly diminishes the cross-task generalization ability of VProg. Besides, the non-differentiable nature of VProg prevents direct fine-tuning on specific VR tasks for further performance improvement. Attempt to address these issues, we propose \textbf{SDVP}, a \textbf{S}tepwise \textbf{D}istillation learning strategy for non-differentiable \textbf{V}\textbf{P}org across various VR tasks. Specifically, our SDVP stepwise distills the capabilities of existing, well-trained small task-specific models for decomposed visual sub-tasks in VProg into the much larger VLMs invoked by corresponding visual sub-modules. Besides, distilling the knowledge of little-size task-specific models into pre-trained larger VLMs rather than replacing them helps keep the cross-task abilities of VProgs.
Extensive and comprehensive experimental results on different VProg frameworks
demonstrate that our SDVP obtains significant performance gains on specific VR benchmarks, i.e., GQA (+2.4\%) and NLVRv2 (+6.2\%) for VisProg and GQA (+6.5\%) and NLVRv2 (+4.0\%) for ViperGPT, and also maintains a promising performance for VProg on unseen and previous VR tasks. 
\end{abstract}


\section{Introduction}
\IEEEPARstart{V}{isual} reasoning (VR) tasks~\cite{johnson2017clevr, hudson2019gqa, nlvr_dataset,nlvr} usually represent a category of complex Visual Question Answering (VQA) tasks~\cite{antol2015vqa,krishna2017visual,goyal2017making} that require multi-step reasoning for humans to solve. Methods for solving VR tasks can be divided into two kinds. One kind is the end-to-end single-step processing, which includes general multi-modal large models~\cite{blip, li2023blip2, zhu2024minigpt, wang2023image, liu2024visual} and task-specific  models~\cite{nguyen2022cvpr_cfr_coarse, anderson2018bottom}, all of which are purely deep neural networks. Such methods can achieve notable performance but usually lack interpretability. The other kind is the neuro-symbolic method~\cite{yi2018nips_neural, mao2019neuro, amizadeh2020neuro}, which employs multi-step reasoning to answer VR questions and provide stronger interpretability. However, when only the answers to VR questions serve as supervision, neuro-symbolic methods have to optimize both perception modules which usually utilize neural networks, and the symbolic formula representation of the reasoning structure, leading to low optimization efficiency~\cite{mao2019neuro,yi2018nips_neural
}. This is particularly challenging when addressing complex visual reasoning tasks in real-world scenarios, where many existing works rely on existing symbolic reasoning programs from datasets as supervision to learn the reasoning structure~\cite{amizadeh2020neuro}, which incurs high acquisition costs~\cite{hudson2019gqa}.


\begin{figure}
\centering
\includegraphics[width=1\linewidth]{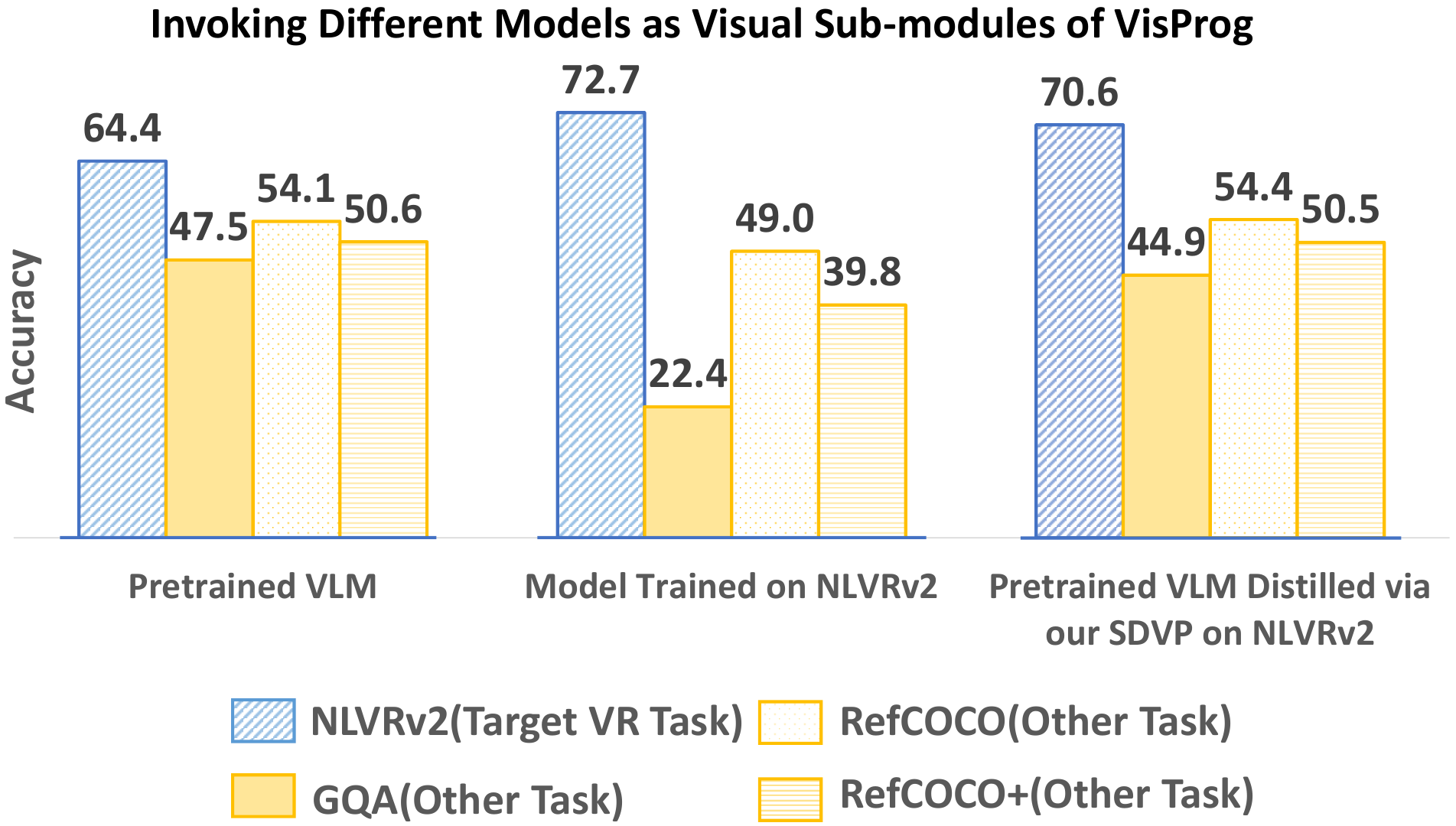}
\vspace{-15pt}
\caption{Performance analysis of VisProg~\cite{Gupta2022cvpr_VisProg}, a classical VProg framework, to invoke different visual models for the four VR tasks. Note that, the target VR task is on NLVRv2~\cite{nlvr}. Invoking the existing models trained on NLVRv2 to perform as the relevant visual sub-modules enables VisProg to perform superior on NLVRv2, but the capabilities of VisProg on other tasks significantly decline. Our SDVP, which distills the capabilities of models well-trained on NLVRv2 into the sub-modules of VisProg, allows VisProg to improve performance on NLVRv2, while essentially maintaining its capabilities on other tasks, thus preserving the cross-task generalization ability of VisProg.
}
\vspace{-15pt}
\label{fig1}
\end{figure}

The emerging VProg~\cite{Gupta2022cvpr_VisProg,suris2023vipergpt} utilizes the built-in knowledge of large language models (LLMs)~\cite{gpt3, codellama, codex, li2023starcoder, nijkamp2023codegen,luo2024wizardcoder}, to directly generate high-quality visual programs from textual questions in VR tasks. As shown in the left part of Fig.~\ref{fig2}, 
The produced visual program is then executed by invoking pre-trained visual models and Python built-in functions, and finally, outputting the answer to the original questions. VProg can also be seen as a form of generalized neuro-symbolic method. Unlike previous neuro-symbolic methods, VProg leverages the knowledge in LLM to generate reasoning structures, i.e. visual programs, in a zero-shot setting. Since VProg uses pre-trained visual models~\cite{weissenborn2022simple, blip,li2023blip2,li2022cvpr_grounded} for its perception modules, it can process VR tasks without training. Furthermore, by decomposing the original VR task into visual sub-tasks using generated visual programs, VProg achieves cross-task generalization ability. It can handle various VR tasks by utilizing a set of visual models tackling visual sub-tasks decomposed from dynamically generated visual programs. Despite these advantages of interpretability, no need for training, and cross-task generalization ability, 
previous works~\cite{Gupta2022cvpr_VisProg,suris2023vipergpt} show that the accuracy of VProg on typical VR tasks (such as GQA and NLVRv2) is significantly lower than well-trained task-specific models.
Enhancing performance, without compromising interpretability and cross-task generalization ability, is a crucial requirement for increasing the practical utility of VProg on VR tasks.

Previous efforts for enhancing the performance of VProg on specific VR tasks focus on improving the quality of visual programs produced by LLM~\cite{khan2024self}. However, error analysis in ~\cite{Gupta2022cvpr_VisProg} reveals that for VProg, on VR tasks, the proportion of errors caused by visual sub-modules is higher than that of code errors in visual programs. Therefore, increased efforts should also concentrate on boosting visual sub-modules of VProg to perform better on target VR tasks, while concurrently preserving their original proficiency in other tasks for maintaining cross-task generalization ability. 

As an agent framework, one straightforward way to enhance the capabilities of VProg sub-modules on the specific VR task is to employ visual models more powerful on that task. As shown in Fig.~\ref{fig1}, when invoking models specifically trained on the target VR task as visual sub-modules of VProg, it performs well on that task but has poor generalization performances on other tasks, since these task-specific models are only trained on the target VR task. When using the pre-trained large Visual-Language Models (VLMs) as the visual sub-modules, it shows reasonable performance across multiple tasks but usually does not perform as well on the target VR task compared to invoking the task-specific models. Our experiments demonstrate that this remains the case even when employing recent powerful VLMs (see Table~\ref{tab: invoking_VLMs}). 
Although the growing VLMs have consistently improved these visual sub-modules of VProg, it would be beneficial and more cost-effective if VProg could be fine-tuned for the target VR task while maintaining the ability for other tasks.

However, fine-tuning VProg on VR tasks is quite challenging due to the non-differentiable nature of VProg as a whole~\cite{khan2024self}. As shown on the left side of Fig.~\ref{fig2}, although the visual sub-modules within VProg, such as `find' `verify\_property', and `simple\_query', invoke internally differentiable visual models, the visual program discretizes and truncates the output from the neural networks before storing a specific result instead of a distribution in variables. Moreover, subsequent non-differentiable Python operations are usually performed on these variables. Together, these two factors render the entire VProg framework non-differentiable. This non-differentiable property impedes the direct fine-tuning of VProg on specific VR tasks.

Attempting to overcome the above difficulty, we creatively propose SDVP: a Stepwise Distillation learning strategy for non-differentiable VProg on specific VR tasks. Our motivation is as follows: although the VProg is non-differentiable as a whole, the learnable part within it - the invoked visual models are differentiable. Therefore, we can design a step-by-step learning strategy, optimizing the visual models invoked by VProg separately to overcome the non-differentiability issue. However, this requires that these visual sub-modules have corresponding labels, i.e., VProg needs execution process labels. Unfortunately, VProg lacks such process labels in VR tasks. Considering that distillation methods can address the lack of labels by providing pseudo-labels~\cite{hinton2015distilling}, we can have VProg invoke proprietary models that have been trained end-to-end on the target VR tasks to generate process pseudo-labels, and subsequently distill these into our pre-trained VLM. As analyzed earlier, since VProg invoking these proprietary models performs well on the target tasks, we can infer that they are capable of generating high-quality VProg process pseudo-labels for distillation. Moreover, unlike traditional end-to-end distillation, our approach involves stepwise distillation along the visual programming steps of VProg, effectively decoupling the original VR task into sub-visual tasks. This method not only enhances the performance of the pre-trained VLM on the target VR tasks but also effectively prevents catastrophic forgetting on other tasks, maintaining multitasking generality. Our experiments validate the aforementioned points, and further mechanistic analysis is provided in Sec.~\ref{subsec: alleviate forgetting}.

Specifically, 
our SDVP proposes to select the differentiable visual sub-modules from VProg, 
then adopt a stepwise distillation strategy to fine-tune these modules with pseudo-labels provided by existing proprietary models well-trained on the target VR task. In this way, our SDVP can overcome the non-differentiable limitation of VProg, and achieve superior performance over the original VProg on the target VR task. 
Additionally, our experimental results show that when the VLM for the selected visual sub-module is trained on specific VR tasks using our SDVP, it effectively maintains the multi-task generality of VProg. That is, VProg experiences little performance degradation on other tasks or on tasks previously learned through SDVP, unlike traditional end-to-end task learning strategy for neural network models which often suffer ``Catastrophic Forgetting" after trained on a new task. We believe that one fundamental reason for the ``Catastrophic Forgetting" in end-to-end neural networks is the significant gap between the input distributions of new tasks and old tasks. VProg, by decoupling the original visual tasks into multiple visual sub-tasks through the LLM-generated visual program, thereby narrowing the distribution gap between the same visual sub-tasks corresponding to different original visual tasks. 
This enables using our SDVP to learn the decoupled visual sub-tasks of new VR tasks, and then using VProg for inference can effectively alleviate the forgetting of pre-trained capabilities or previously acquired skills on previously learned VR tasks.
In other words, we have found that learning and inference strategies based on task-decomposed frameworks like VProg have a certain degree of anti-forgetting potential during continuous learning processes. For specifics, please refer to our experimental results (Tab.~\ref{tab:SDVP_GQA} and~\ref{tab:SDVP_GQA_NLVRv2}). Additionally, we provide a more detailed analysis and design additional comparative experiments to illustrate this point (Sec.~\ref{subsec: alleviate forgetting}).

 \textbf{Our main contributions are listed as follows:} 

i) To the best of our knowledge, we are the first ones to enable VProg fine-tuning on specific VR tasks by proposing a cost-effective Stepwise Distillation strategy SDVP. Our SDVP can address the learning challenge introduced by the non-differentiability inherent in VProg by performing stepwise distillation to provide process pseudo-labels of the objective VR tasks and then distilling them into the pre-trained VLMs invoked by VProg, enabling this agent framework learning on the target VR tasks.


ii) Our SDVP proposes to use proprietary small models as teachers to distill their capabilities into larger, general-purpose models. Our SDVP is distinct from traditional distillation practices, which typically involve distilling the capabilities of larger, more powerful models into smaller models. 

iii) Our SDVP not only enhances performance on the target VR task but also effectively maintains the multi-tasking generality of the VProg framework. We think the reason for this is that Stepwise Distillation under VProg decouples the original VR task into visual sub-tasks and the input of those visual sub-tasks is the interesting image area instead of the whole image, which narrows the input gap between different original VR tasks.

Extensive and comprehensive experimental results on different VProg frameworks demonstrate that our SDVP obtains significant performance gains on target VR tasks and also maintains a promising performance for VProg on unseen and previous VR tasks.

\section{Related Work}

\subsection{VProg based on Large Language Models}

VProg~\cite{Gupta2022cvpr_VisProg, suris2023vipergpt} can be regarded as a new version of the neuro-symbolic approach for dealing with composite visual tasks. It utilizes Large Language Models (LLMs)~\cite{gpt3, codellama, codex} to generate and execute visual programs, to handle specific composite visual tasks. The execution process often includes pre-defined visual sub-modules invoking corresponding pre-trained VLMs~\cite{weissenborn2022simple, li2022cvpr_grounded, blip, li2023blip2} and various functions from Python libraries (e.g., sorting, mathematical operations, etc.). Since the performance of VProg on VR tasks is not good enough~\cite{Gupta2022cvpr_VisProg, suris2023vipergpt}, we attempt to find various ways to improve it. However, VProg is non-differentiable~\cite{khan2024self}, so it cannot be fine-tuned on specific tasks directly. In addition, although the code generation quality of LLMs has been improving with each version iteration, our analysis reveals that the main errors of VProg originate from its visual sub-modules (Sec.~\ref{subsec: error_analysis}). These errors cannot be rectified through improvements in the quality of generated visual programs. Therefore, enhancing the capabilities of the visual sub-modules is a requirement for improving the performance of VProg. Considering this, our SDVP utilizes an existing well-trained model on the target VR task to teach the invoked visual sub-modules to help VProg perform better on that task.

\subsection{Knowledge Distillation}


Knowledge distillation~\cite{gou2021knowledge,wang2021knowledge} is a method that distills the capabilities of a teacher model into a student model by having the student model mimic the output of the teacher model given the same input. This method was first proposed by Hinton et al.~\cite{hinton2015distilling}. In previous studies, knowledge distillation has generally served several purposes. i) Model compression~\cite{hong2022analysis, sun2019patient,li2020few,wang2019private,aguinaldo2019compressing}: knowledge is distilled from one model into another model with significantly fewer parameters; ii) Enhancing accuracy~\cite{prakosa2021improving}: in this case, the student model learns the soft label of the teacher model under the same input of the training set, often achieving better generalization than directly learning the hard label, which results in higher accuracy on the test set;
iii) Domain transfer~\cite{fang2021mosaicking, belal2021knowledge,niu2022respecting}: this typically involves using the soft label of the teacher model trained on the source domain on the input of the target domain, distilled into the student model learning on the target domain, aiding the student model in gaining better generalization capabilities on the target domain when labels are lacking.

Our SDVP is a form of knowledge distillation for domain transfer, differing from prior studies by using the original question (or statement) and image data from VR tasks as the source domain, and the cropped image and decoupled sub-questions as the target domain for visual sub-modules in VProg. 
Prior knowledge distillation research focuses on distilling knowledge from a large teacher model into a smaller student model. This includes recent works~\cite{hsieh2023distilling, li2023symbolic, wang2024can} where knowledge of an LLM with huge parameters is distilled into another LLM with much fewer parameters step by step along with the chain of thought (COT)~\cite{wei2022nips_cot_chain}. ~\cite{hu2024visual} considers distillate the knowledge of a VProg into one large VLM to improve the performance of the VLM. Our SDVP, however, stepwisely distills knowledge from a small-scale task-proprietary visual model into larger pre-trained visual models invoked as the visual sub-modules of VProg, enhancing the performance of VProg on target VR tasks while trying to maintain the ability on other visual tasks.

\subsection{Catastrophic Forgetting}

``Catastrophic Forgetting"~\cite{wang2024comprehensive}, 
is a substantial challenge in continual learning. It generally manifests in two forms: one is a significant performance degradation on old tasks when proprietary models are trained on new tasks~\cite{yao2019adversarial}; the other is a substantial decrease in the versatility of general-purpose models, after fine-tuning on specific task data, despite extensive pre-training. 
Solutions to ``Catastrophic Forgetting" are typically categorized into three types: dynamic networks~\cite{amer2019reducing}, gradient regularization~\cite{kirkpatrick2017overcoming}, and data replay strategies~\cite{hayes2020remind, lesort2020continual}.

VProg usually invokes large-size pre-trained multimodal models(e.g., BLIP-2~\cite{li2023blip2}, InternVL~\cite{chen2024internvl}, etc.) to perform visual sub-modules. It faces the second kind of ``Catastrophic Forgetting" if fine-tuning a specific VR task. 
In our SDVP, we employ a stepwise distillation strategy to leverage the narrowed input domain gap of the same visual sub-modules of VProg across different VR tasks to alleviate the forgetting of versatility while fine-tuning the target VR task.

\section{Methodology}
This section introduces our SDVP method. The first subsection will formalize the VProg framework~\cite{Gupta2022cvpr_VisProg, suris2023vipergpt} and the tasks we aim to address. The second subsection will detail the various components of the SDVP method, including the use of Adapters for decoupled sub-tasks aligning interfaces with the task-specific teacher model, the mechanism for generating pseudo-labels on intermediate tasks, and how to conduct stepwise distillation. 

\subsection{Formalization of VProg on VR Tasks}
Our SDVP aims to fine-tune the non-differentiable VProg on specific VR tasks while maintaining the multi-tasking generality of VProg. 
This section introduces VR tasks and the formal description of handling VR tasks using VProg.

For a VR task $\tau$ with the question and image pair $<q, I>$, $q$ represents the question of one sample in $\tau$, and $I$ represents the corresponding image. For VProg, a visual program $P$ to solve $q$ is first generated under prompt $pr$ through an LLM:
\begin{equation}
    P = f(pr,q; \theta_{llm})
\end{equation}
in which $f(\cdot; \theta_{llm})$ means a LLM model with parameter $\theta_{llm}$. The generated visual program $P$ often contains multiple steps, decompose Task $\tau$ into $N_P$ visual sub-tasks $\tau=\{\tau_j\}^{N_P}_{j=1}$. 
$P_k$ denotes the $k$-th line of $P$. 
Let us assume that $P_k$ processes the $j$-th visual sub-task $\tau_j$ decomposed from Task $\tau$, and $\tau_j$ with $I_j$ as the image input, $T_j$ as the text input. Executing $P_k$ to get the prediction $y_j$ for $\tau_j$:
\begin{equation}
\begin{split}
\label{eq:sub_y_Ti}
    y_j  
                      &= \mathbf{Exe}[P_k(I_j, T_j)] \\
                      &= f(I_j, T_j; \theta_k)
\end{split}
\end{equation}
where $\mathbf{Exe}(\cdot)$ is a program excution operation. Eq.~\ref{eq:sub_y_Ti} means executing $P_k$ with $(I_j, T_j)$ as input parameters by invoking a visual model with $\theta_k$ as weight.
$(I_j, T_j)$ is the input of sub-task $\tau_j$, $T_j$ is produced along with the production of visual program $P$; $I_j$ is 
usually a part of the original image $I$, and is generated by the previous visual sub-task in $P$ through:
\begin{equation}
\begin{split}
    I_j &= \mathbf{crop\_op}[\mathbf{Exe}[\mathop{P_l}\limits_{l<k}(I_l, T_l)]]\\
                        &= \mathbf{crop\_op}[f(I_l, T_l; \theta_{loc})]\\
\end{split}
\end{equation}
where $f(I_l, T_l; \theta_{loc})$ is an bounding-box output of an object detection model with $(I_l, T_l)$ as input and $\theta_{loc}$ as weight parameter. This detection model is invoked by $P_l$ as the object location module to locate the object $T_l$ from the image $I_l$. $\mathbf{crop\_op}$ is one operation of the crop operation set: 
$\{\mathbf{crop, crop\_above, crop\_below, crop\_left, crop\_right}\}$ for cropping the original image.
\noindent It is specifically determined by the next program line of $P_l$, representing cropping the target area or one part of its surrounding areas. By executing the entire visual program $P$, we get the answer prediction for the sample under the target task $\tau$:
\begin{equation}
    y = \mathbf{Exe}[P(I; \Theta)]
\end{equation}
where $\Theta$ means weight parameters of all visual sub-modules invoked by visual program $P$.
The objective of our methodology is to enhance the estimation of $y$ for Task $\tau$
, while concurrently minimizing the detriment to $y'$ for other task $\tau'$. We propose SDVP to chase this objective.

\subsection{Stepwise Distillation for VProg}
\subsubsection{Motivation}

Our goal is to enable VProg to learn specific VR tasks to enhance its performance. However, a major challenge is the non-differentiability between the multiple steps in VProg, which impedes the direct use of gradient-based optimization methods using the original VR task labels to optimize the framework. While reinforcement learning could address this issue, considering its generally low optimization efficiency, we seek a more convenient learning strategy. We note that the non-differentiability in VProg primarily occurs between steps, while the internal operations of visual subtasks within these steps are differentiable. Therefore, if we had labels for these intermediate visual subtasks, we could supervise and optimize the visual models invoked at each step of VProg. Unfortunately, we do not have these intermediate process labels.
However, we have many neural network models that have been end-to-end trained on these VR tasks. Noticing that VProg decouples complex VR questions into multiple simpler VQA tasks, we think that the capabilities of these end-to-end models can generalize to these decoupled, simpler VQA tasks. Consequently, we propose using the existing end-to-end neural network that is well-trained on the target VR task as the teacher (usually smaller models). VProg invokes the teacher to collect its outputs on the decoupled visual subtasks as pseudo-labels for stepwise optimization of the pre-trained VLMs invoked by VProg. Therefore, we introduce a stepwise distillation learning strategy for VProg on VR tasks. 

Besides, cause VProg selects the interesting part of the original image and decouples the complex long original question into several simple questions, it reduces the input domain gap of the same visual sub-modules between different original visual tasks. This is somewhat similar to methane and ethane being two different molecules, which, when broken down into atoms, become the same, consisting of carbon and hydrogen atoms. A much smaller input domain gap alleviates the forgetting of previous abilities while fine-tuning new target VR tasks. For relative experimental results please refer to Tab.~\ref{tab:SDVP_GQA} and Tab.~\ref{tab:SDVP_GQA_NLVRv2}, further relative analyses please refer to Sec.~\ref{subsec: alleviate forgetting}.

\begin{figure*}[t]
\centering
\includegraphics[width=1\textwidth]{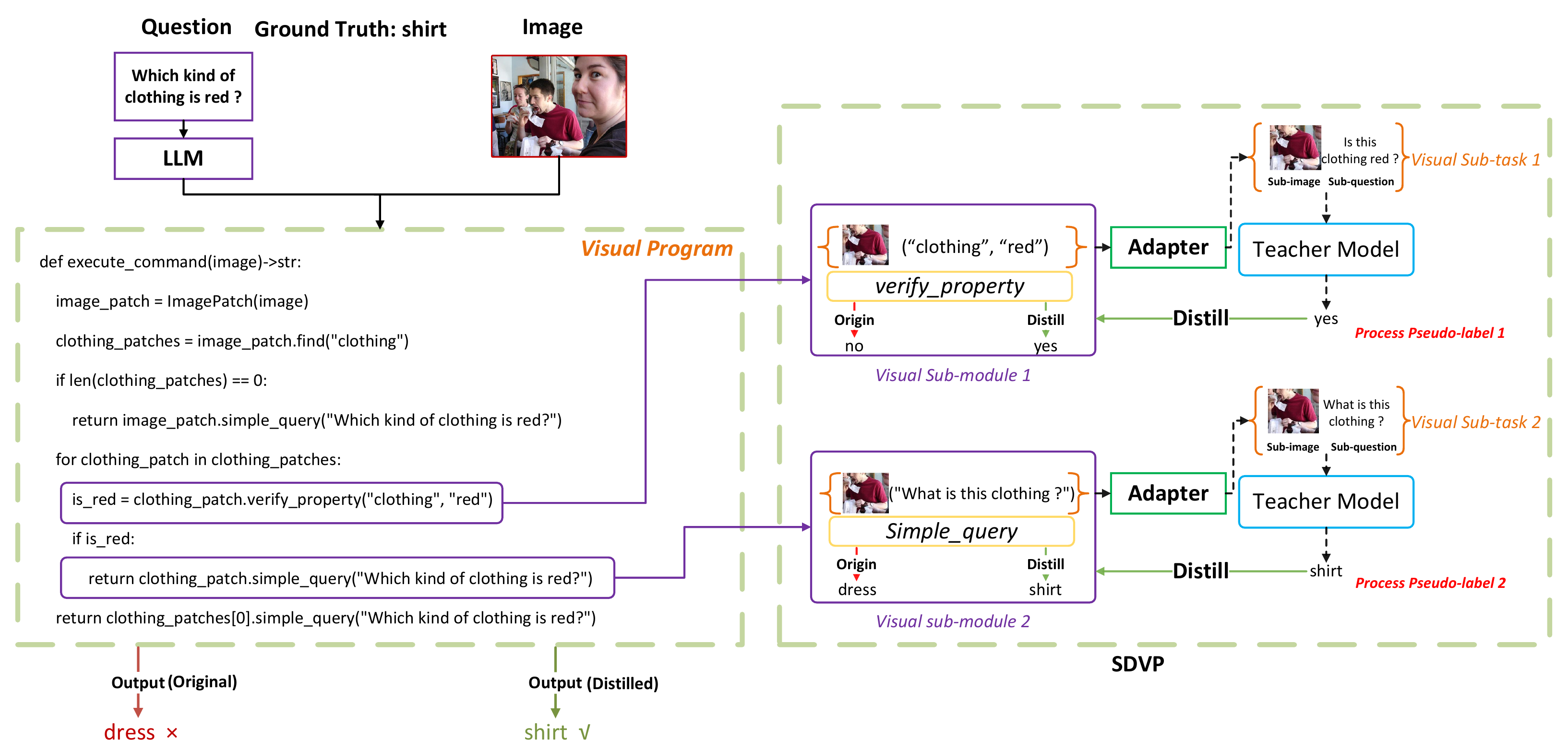}
\vspace{-15pt}
\caption{The overview of our proposed SDVP pipeline. SDVP aims to fine-tune visual sub-modules to produce the correct predictions for visual sub-tasks, so as to access the correct final answer of the original question. The right part displays how our SDVP works. During the execution of the program, the visual sub-module ``verify\_property" is called to verify whether the clothing in the image is red, and it answers ``no" wrongly. To rectify it, we use the teacher model to answer the same question, then we let ``verify\_property" learn it through distillation. After distillation, ``verify\_property" gives a correct answer ``yes". Similarly, after distillation the visual sub-module ``simple\_query" would answer ``shirt" correctly instead of ``dress". Because of the correct reasoning of each sub-module, in the final step, we could get the correct answer ``shirt" to the original question ``Which kind of clothing is red ?" with the given image.  }
\vspace{-15pt}
\label{fig2}
\end{figure*}

As shown in Fig.~\ref{fig2}, Our SDVP includes three parts: an ``Adapter" that transforms the inputs of the visual sub-modules into a format acceptable by the teacher model, where the teacher model refers to the task-specific model already available in the community, which has been trained with outcome supervision; a ``Pseudo-label Production" part, which produces supervision for the execution results of the visual sub-modules to be used in stepwise distillation; a ``Stepwise Distillation" part, which distills the process knowledge from existing task-specific teacher model into visual sub-modules in VProg step by step.

\subsubsection{Adapter}
Our SDVP uses the existing task-specific model $f({\cdot};\theta_{te})$ as the teacher to generate supervision for each visual sub-task $\tau_j$ decomposed from the original VR task $\tau$. But the interface of $\tau_j$ in $P_k$ is usually different with $f({\cdot};\theta_{te})$ for the original target VR task. Here, we use an LLM $f({\cdot};\theta_{llm})$ as the Adapter to transform the code and input parameters of $P_k$ into acceptable inputs of $f({\cdot};\theta_{te})$:
\begin{equation}
    q_j = f(pr^*, P_k(I_j, T_j);\theta_{llm})
\end{equation}
in which $q_j$ is a sub-question, and $pr^*$ is the prompt manually designed for the Adapter of Task $\tau$.

\subsubsection{Pseudo-Label Generation}
Upon acquiring inputs that the teacher model $f({\cdot};\theta_{te})$ can accept through the Adapter, we feed them into the teacher model and use the output from the teacher model as the pseudo-labels $y^{te}_j$ for that particular visual sub-task $\tau_j$. The specific process is as follows:
\begin{equation}
\begin{split}
    y^{te}_j 
                               &= f((I_j, q_j); \theta_{te}).
\end{split}
\end{equation}
The teacher model $f({\cdot};\theta_{te})$, whose weights can be conveniently obtained from a public repository, is trained on the target VR task $\tau$ using the original input data $<q, I>$ and the ground truth $y_{gt}$. The value of the process pseudo-labels $y^{te}_j$ generated hinges on the premise that $f({\cdot};\theta_{te})$ can generalize effectively to $(I_j, q_j)$. Due to the combinatorial nature of VR task questions and the variability in length within the dataset of Task $\tau$, $q_j$ transformed by our Adapter closely resembles the shorter questions $q$ in Task $\tau$. Consequently, we speculate that the teacher model can generalize well to the intermediate sub-tasks within VProg on the target VR task, thereby producing high-quality process pseudo-labels. Experiment results in Sec.~\ref{subsec: exps_results} demonstrate our hypothesis.

\subsubsection{Stepwise Distillation}

To enhance the prediction $y$ on Task $\tau$ predicted by VProg, we performed a sampling error analysis (Sec.~\ref{subsec: error_analysis}). This revealed that the principal errors in $y$ stem from incorrect answer predictions $y_j$ for visual sub-tasks in Eq. ~\ref{eq:sub_y_Ti}. We consider that the primary reason is the inability of the pre-trained visual sub-module $f({\cdot}; \theta_k)$, employed by VProg to execute the corresponding visual sub-task $\tau_j$, 
to effectively generalize to the unseen inputs $(I_j, T_j)$ in Eq.~\ref{eq:sub_y_Ti}.
This leads to the goal of how to optimize $y_j$ without the corresponding ground-truth labels $y^{gt}_j$. 
As analyzed in the last paragraph, task-specific models for the target VR task can generalize well to the corresponding visual sub-tasks decomposed by VProg. However, merely using these task-specific models to replace the pre-trained models originally invoked by VProg to act as corresponding visual sub-modules would lead to a drastic decline in performance on other tasks, severely impairing the cross-task generalization ability of VProg. 

Although directly invoking the task-specific model would cause VProg to lose its cross-task generalization ability, we can stepwise distill the pseudo-labels generated by the task-specific model on visual sub-tasks into the corresponding visual sub-modules. This enhances the performance of these visual sub-modules on corresponding visual sub-tasks, thereby better adapting VProg to the target VR task. To maintain the cross-task generalization ability of VProg, it is crucial to prevent ``Catastrophic Forgetting" during the distillation process. From past experience, ``Catastrophic Forgetting" typically occurs due to a large distribution gap between new and old task data. 
We find that due to the task decomposition operation of VProg, the gap between inputs of the same type of visual sub-tasks derived from different VR tasks is much smaller than that of the original VR tasks (more analyses can be found in 
the supplementary material). Therefore, we believe that learning on these decomposed visual sub-tasks has better anti-forgetting characteristics.

Considering these, in our SDVP, we propose a stepwise distillation strategy to possess a stronger resistance to forgetting while fine-tuning the target VR task. 
Please refer to Sec.~\ref{subsec: exps_results} for relevant experimental results. 
We will next describe the specific operations for stepwise distillation for VProg on Task $\tau$.

After getting $y^{te}_j$, we utilize it as the pseudo-label for $y_j$, and employ cross-entropy loss

\begin{equation}
\label{eq:SWD_loss}
    L_{SDVP} = -\sum_{j}^{N_p}y^{te}_j\log(y_j) = -\sum_{j}^{N_p}y^{te}_j\log[f(I_j, T_j; \theta_k)]
\end{equation} 
to supervise the optimization of the visual sub-module $f({\cdot}; \theta_k)$ that VProg invoked in $P_k$ to enhance performance on visual sub-task $\tau_j$. After executing this distillation process step by step along with the Visual Program $P$, we can finish the fine-tuning process of non-differentiable VProg on the target VR task $\tau$.

\section{Experiments}
\subsection{Dataset and Experimental Setting}
We sequentially distill the knowledge of existing teacher models well-trained on GQA~\cite{hudson2019gqa} and NLVRv2~\cite{nlvr} into the visual sub-modules under two VProg frameworks: VisProg~\cite{Gupta2022cvpr_VisProg} and ViperGPT~\cite{suris2023vipergpt}. We assess the performance of SDVP on GQA, NLVRv2, and Visual Grounding (VG~\cite{vg}) tasks after each round of distillation, to evaluate and compare the performance enhancement of SDVP on the currently focused VR task, and whether this round of distillation has impaired the performance of VProg on other tasks.

GQA~\cite{hudson2019gqa} is a VR task dataset based on real-world scenario question answering. To control the costs of getting LLM-produced visual programs, we have randomly drawn 140,000 and 80,000 samples (excluding labels) from the GQA train-balanced split for our SDVP to stepwisely distill invoked visual models of VisProg and ViperGPT respectively, maintaining the question type proportionality of those samples. We evaluate the performance of our SDVP for those two VProg frameworks in the test-dev balanced split of GQA.
NLVRv2\cite{nlvr} is another VR task dataset with two real-world images and a statement (waiting to judge `True' or `False') as the input. We use the whole train set data (excluding labels) for distillation and evaluate our SDVP in the public test set split of NLVRv2. 
As for VG tasks, we adopt the testA split of RefCOCO~\cite{kazemzadeh2014refcoco} and RefCOCO+ as the evaluation dataset split. 


In our SDVP, we respectively choose CFR~\cite{nguyen2022cvpr_cfr_coarse} and Beit3-NLVRv2~\cite{wang2023image} (means Beit3 fine-tuned on NLVRv2) as the teacher model for GQA and NLVRv2. The student models are the visual models invoked to perform as the ``VQA'' module in VisProg and the ``verify\_property" ``best\_text\_match" and ``simple\_query" modules in ViperGPT. Those three modules in ViperGPT invoke just the same one pre-trained VLM. We evaluate our SDVP under the situation in which the pre-trained VLMs BLIP~\cite{blip} and InternVL-1.5~\cite{chen2024fargpt4vclosinggap} are respectively invoked to perform as the abovementioned modules. 
The visual programs for VisProg on GQA are produced by codellama-python-34B~\cite{codellama}, on RefCOCO, RefCOCO+, and NLVRv2 are produced by GPT-3.5-turbo. And the visual programs for ViperGPT on GQA, RefCOCO, RefCOCO+, and NLVRv2 are produced by GPT-3.5-turbo. 

\subsection{Implementation Details}
\subsubsection{\textbf{Interface Alignment on NLVRv2}}
We employ stepwise distillation to rectify the visual sub-modules in VProg with the sub-questions and the pseudo-labels given by the teacher model~\cite{nguyen2022cvpr_cfr_coarse, wang2023image}. However, for the sub-questions of NLVRv2\cite{nlvr}, we cannot feed them to the teacher model directly, because there exists a gap between the inputs of visual sub-modules (an image and a question) and the teacher model (two images and a statement). In order to get the pseudo-labels, We need to achieve the interface alignment. 

Since each input set of the NLVRv2 teacher model is a statement and two images, we need to convert the sub-questions within VProg into corresponding statements, which can be finished by LLM (Here we use GPT-3.5-turbo). First of all, we categorize these sub-questions into four types: ``How many", ``Is there/Are there", ``Is/Are/Do/Does" and ``other" according to the first two words of each sub-question. Since the answers to the questions ``Is there/Are there" and ``Is/Are/Do/Does" are only ``yes" or ``no", we convert these two types of sub-questions into corresponding statements directly, collect the ``True" or ``False" answers from the Teacher, then change those answers into ``yes" or ``no" to get the pseudo-labels for those sub-questions. As for ``How many" type questions, we change them into ten statements with the included number different from 0 to 9. Then we choose the number corresponding to the highest probability of ``True" or the lowest probability of ``False" (if no answer is ``True") from the Teacher to produce the pseudo-labels for this type of sub-questions. For ``other" type sub-questions, we discard them directly because they are few.

Cause the teacher model for NLVRv2 expects to accept two images, we just copy the sub-image corresponding to the sub-questions and add ``in the left image" or ``in the right image" randomly in the constructed statements (also utilizing GPT-3.5-turbo). 



\subsubsection{\textbf{More Details for Experimental Results}}
For ViperGPT, we performed stepwise distillation on three visual sub-modules: ``verify\_property", ``best\_text\_match", and ``simple\_query". In~\cite{suris2023vipergpt}, these three interfaces respectively invoked the neural network models XVLM~\cite{zeng2022icml_xvlm_multi}, XVLM, and BLIP-2~\cite{li2023blip2}. 
In our experiments, the neural network model used for all three sub-modules was the same pre-trained VLM, and our experiments confirmed that this change had minimal impact on the performance of ViperGPT on the tasks. The reason the VLMs like BLIP could be used to implement the ``verify\_property" and ``best\_text\_match" modules is that our SDVP includes an Adapter capable of transforming the interface input format, thus converting the text inputs of these two interfaces also into sub-questions for the VLMs. We take this setting for convenience, our SDVP can also distill multiple student visual models.

\subsection{Evaluation Details}
In our experiments, we adopt Accuracy (ACC) to evaluate the performance on GQA and NLVRv2, and the average Intersection over Union (IoU) to assess the performance on VG tasks.
To assess the changes in the capability of VProg on other tasks after learning about the target VR task using our SDVP, we test the performance of VProg on all four tasks before and after each distillation session.

\subsection{Performance of VProg while invoking different pre-trained VLMs}

\begin{table*}[h]
    \caption{Results of VProg while invoking various recently powerful pre-trained VLMs under zero-shot setting. All results are obtained under our implementation.
    }
\footnotesize
\centering
    \begin{tabular}{cc|cccc}
    \hline
    \multirow{2}{*}{\textbf{VProg\_framework}} & \multirow{2}{*}{\textbf{Invoked Visual Model}} & \textbf{GQA} & RefCOCO & RefCOCO+ & \textbf{NLVRv2} \\
     &  & test-dev balanced & testA & testA & public test \\
     \hline
     & & Accuracy(\%) & \multicolumn{2}{c}{IoU} & Accuracy(\%) \\
     \hline
     \multirow{6}{*}{VisProg~\cite{Gupta2022cvpr_VisProg}} & BLIP~\cite{blip} & 45.1 & 55.0 & 51.2 & 65.7 \\
     & InstructBLIP~\cite{dai2023instructblipgeneralpurposevisionlanguagemodels} & 45.0 & 46.6 & 44.2 & 68.2 \\
     & InternVL-1.5~\cite{chen2024fargpt4vclosinggap} & 51.0 & 49.0 & 47.1 & 68.9\\
     & InternVL-2 & 51.1 & 49.2 & 46.9 & 68.2 \\
     & CFR~\cite{nguyen2022cvpr_cfr_coarse} (Teacher) & \textbf{60.3} & 44.5 & 35.0 & 45.0 \\
     & Beit3-NLVRv2~\cite{wang2023image} (Teacher) & 22.4 & 49.0 & 39.8 & \textbf{72.7} \\
      \hline
      \multirow{6}{*}{ViperGPT~\cite{suris2023vipergpt}} & BLIP & 43.1 & 58.8 & 50.8 & 60.5\\
      & InstructBLIP & 48.5 & 60.3 & 52.0 & 67.6\\
     & InternVL-1.5 & 54.0 & 61.7 & 56.2 & 67.1 \\
     & InternVL-2 & 53.5 & 61.9 & 56.7 & 67.5\\
     & CFR (Teacher) & \textbf{55.9} & 59.7 & 52.6 & 59.8 \\
     & Beit3-NLVRv2 (Teacher) & 33.2 & 59.7 & 51.2 & \textbf{70.4} \\
      \hline
    \end{tabular}
    \label{tab: invoking_VLMs}
\end{table*}

As an agent framework, one way to improve the performance of VProg is to invoke more powerful visual models. To analyze if there is room for our SDVP to improve VProg while VProg invokes more powerful visual models. we have two VProg frameworks, VisProg and ViperGPT, each invokes a series of different visual models ranging from BLIP to InternVL-2, one of the most powerful and recent open-source visual models. As a comparison, we also make VProg respectively invoke two teacher models, CFR (for GQA) and Beit3-NLVRv2 (for NLVRv2), as the corresponding visual sub-modules. we test the performance of VProg under four tasks in each scenario and record the results in Tab.~\ref{tab: invoking_VLMs}. From Tab.~\ref{tab: invoking_VLMs}, it can be seen that although invoking the recent state-of-the-art VLMs, the performance of VProg on the target VR task still significantly lags behind VProg invoking the teacher model specifically well-trained on the target VR task. However, the performance on non-target tasks achieved by VProg with the teacher model invoked is significantly worse than VProg with pre-trained VLMs invoked. It is worth noting that these powerful pre-trained VLMs have been developed through highly costly, large-scale training on massive datasets. When the performance of VProg on the target VR task is limited due to various reasons (such as insufficient capabilities of pre-trained VLMs or restricted running costs), our SDVP can enhance the performance of VProg on the target VR task cost-effectively and efficiently, while maintaining the multitasking general capability of VProg, even without knowing the task ID.

\subsection{Experimental Results of our SDVP}
\label{subsec: exps_results}
To evaluate the effectiveness of our SDVP in improving the performance of VProg(VisProg~\cite{Gupta2022cvpr_VisProg} and ViperGPT~\cite{suris2023vipergpt}) on target VR tasks, we sequentially distill the visual sub-modules in VProg on GQA and NLVRv2 using our SDVP. After each distilling round, we test the performance of VProg on GQA, VG (RefCOCO, RefCOCO+), and NLVRv2. The results are recorded in Tab.~\ref{tab:SDVP_GQA} and Tab.~\ref{tab:SDVP_GQA_NLVRv2} respectively.

\begin{table*}[h]
     \caption{Results for our SDVP on GQA. CFR under Visprog as Teacher represents VisProg with the ``VQA" module invoking the teacher model CFR. CFR under ViperGPT as Teacher represents ViperGPT with the ``verify\_property", ``best\_text\_match", and ``simple\_query" modules invoking CFR. BLIP and InternVL-1.5 under VisProg/ViperGPT represent VisProg/ViperGPT with modules mentioned above respectively invoking the BLIP or InternVL-1.5 model. SDVP$_G$ represents VisProg/ViperGPT with modules mentioned above stepwise distilled on GQA with our SDVP.  
     }
    \footnotesize
\centering
    \begin{tabular}{ccc|cccc}
    \hline
     & \multirow{2}{*}{\textbf{Invoked Visual Model}} & \multirow{2}{*}{\textbf{Method}} & \textbf{GQA}(Target Task) & RefCOCO & RefCOCO+ & NLVRv2 \\
     & & & test-dev balanced & testA & testA & public test \\
     \hline
     & & & Accuracy(\%) & \multicolumn{2}{c}{IoU} & Accuracy(\%) \\
     \hline
     \multirow{2}{*}{Supervised} & & CFR~\cite{nguyen2022cvpr_cfr_coarse} & 72.1 & - & - & - \\
     & & CCO\cite{li2021calibrating} & 56.1 & - & - & - \\
     \hline
     \multirow{2}{*}{Zero-shot}& & PnP-VQA\cite{tiong2022plug} & 42.3 & - & - & - \\
     & & BLIP-2\cite{li2023blip2} & 44.7 & - & - & - \\
     \hline
     \multirow{5}{*}{VisProg} & CFR & Teacher & 60.3 & 44.5 & 35.0 & 45.0 \\
     & \multirow{2}{*}{BLIP~\cite{blip}} & Baseline & 45.1 & 55.0 & 51.2 & 65.7 \\
     & & \textbf{SDVP$_G$ (Ours)} & 47.5(\textbf{+2.4}) & 54.1(-0.9) & 50.6(-0.6) & 64.4(-1.3)\\
     & \multirow{2}{*}{InternVL-1.5~\cite{chen2024fargpt4vclosinggap}} & Baseline & 51.0 & 49.0 & 47.1&68.9 \\
     & & \textbf{SDVP$_G$ (Ours)} & 52.0(\textbf{+1.0}) & 49.1(+0.1) & 46.3(-0.8) & 67.7(-1.2) \\
     \hline
     \multirow{5}{*}{ViperGPT} & CFR & Teacher & 55.9 & 59.7 & 52.6 & 59.8 \\
     & \multirow{2}{*}{BLIP} & Baseline & 43.1 & 58.2 & 50.0 & 60.5 \\
     & & \textbf{SDVP$_G$ (Ours)} & 49.6(\textbf{+6.5}) & 59.7(+1.5) & 53.4(+3.4) & 64.8(+4.3) \\
     & \multirow{2}{*}{InternVL-1.5} & Baseline & 54.0 & 61.7 & 56.2 & 67.1 \\
     & & \textbf{SDVP$_G$ (Ours)} & 54.4(\textbf{+0.4}) & 60.7(-1.0) & 54.7(-1.5) & 66.2(-0.9) \\ 
     \hline
    \end{tabular}
    \label{tab:SDVP_GQA}
\end{table*}

\begin{table*}[h]
    \caption{Results for VisProg and ViperGPT stepwise distilled on NLVRv2 after stepwise distilled on GQA with our SDVP. Here, the baseline is SDVP$_G$ under VisProg/ViperGPT, which have already been distilled on GQA with our SDVP. ViLT-NLVRv2, BLIP-NLVRv2, and Beit3-NLVRv2 mean corresponding networks fine-tuned on NLVRv2; method in the Teacher lines mean using Beit3-NLVRv2 as the relevant visual sub-modules in VisProg/ViperGPT; SDVP$_{GN}$ under VisProg/ViperGPT means corresponding VProg stepwise distilled on NLVRv2 after stepwise distilled on GQA. 
    }
    \footnotesize
\centering
    \begin{tabular}{ccc|cccc}
    \hline
     & \multirow{2}{*}{\textbf{Invoked Visual Model}}& \multirow{2}{*}{\textbf{Method}} & GQA & RefCOCO & RefCOCO+ & \textbf{NLVRv2}(Target Task) \\
     & & & test-dev balanced & testA & testA & public test \\
     \hline
     & & & Accuracy(\%) & \multicolumn{2}{c}{IoU} & Accuracy(\%) \\
     \hline
     \multirow{3}{*}{Supervised}& & ViLT-NLVRv2~\cite{kim2021vilt} & - & - & - & 76.3 \\
     & & BLIP-NLVRv2~\cite{blip} & - & - & - & 83.1 \\
      & & Beit3-NLVRv2~\cite{wang2023image} & - & - & - & 89.4 \\
     \hline
     \multirow{5}{*}{Visprog} & Beit3-NLVRv2 & Teacher & 22.4 & 49.0 & 39.8 & 72.7 \\
     & \multirow{2}{*}{BLIP} & SDVP$_G$ (Baseline) & 47.5 & 54.1 & 50.6 &64.4\\
     & & \textbf{SDVP$_{GN}$ (Ours)} & 44.9(-2.6) & 54.4(+0.3) & 50.5(-0.1) &70.6(\textbf{+6.2})\\
     & \multirow{2}{*}{InternVL-1.5} & SDVP$_G$ (Baseline) & 52.0 & 49.1&46.3 &67.7 \\
     & & \textbf{SDVP$_{GN}$ (Ours)} & 50.4(-1.6) & 50.9(+1.8)&49.4(+3.1) & 71.0(\textbf{+3.3}) \\
     \hline
     \multirow{5}{*}{ViperGPT} & Beit3-NLVRv2 & Teacher & 33.2 & 59.7 & 51.2 & 70.4 \\
     & \multirow{2}{*}{BLIP} & SDVP$_G$ (Baseline) & 49.6 & 59.7 & 53.4 & 64.8\\
     & & \textbf{SDVP$_{GN}$ (Ours)} & 47.4(-2.2) & 61.1(+1.4) & 55.0(+1.6) & 68.8(\textbf{+4.0}) \\
     & \multirow{2}{*}{InternVL-1.5} & SDVP$_G$ (Baseline) & 54.4 & 60.7&54.7 &66.2 \\
     & &\textbf{SDVP$_{GN}$ (Ours)} & 50.7(-3.3) & 61.9 (+1.2)& 56.2(+1.5)&70.0(\textbf{+3.8}) \\
     \hline
    \end{tabular}
    \label{tab:SDVP_GQA_NLVRv2}
\end{table*}

\subsubsection{\textbf{Performance Gain after StepWise Distillation on GQA}}
In Tab.~\ref{tab:SDVP_GQA}, we perform a quantitative evaluation of different methods across several tasks taking GQA as the target task.  
As for the baseline, we respectively make VProg invoking pre-trained BLIP and InternVL-1.5 as the VQA module for VisProg and the ``verify\_property"  ``best\_test\_match" ``simple\_query" modules for ViperGPT.
It can be observed that when those visual sub-modules of VProg invoke the teacher model CFR, although there is significant growth (+15.2\% and +9.3\% for VisProg, +12.8\% and +1.9\% for ViperGPT) on the test-dev balanced split of GQA, the cost is a significant decrease in performance on other tasks, with a maximum decline of 23.9\% compared with VisProg invoking pre-trained InternVL-1.5 on NLVRv2. 
Our SDVP is demonstrated to be beneficial in improving the performance of VProg on the target task GQA without significantly compromising its performance on other tasks. Furthermore, our experiments reveal another interesting phenomenon: the ViperGPT framework with stepwise distillation on GQA not only does not perform worse on NLVRv2 but outperforms the baseline (by 4.3\%). This suggests that our SDVP introduces a transfer learning effect while fine-tuning ViperGPT on GQA. We believe this is owing to the task decoupling of VProg, which reduces the gap between different tasks, resulting in cross-task anti-forgetting effects or even cross-task transfer learning gains. For more details, please refer to Sec.~\ref{subsec: alleviate forgetting}. We believe that the transfer learning effects observed here are due to ViperGPT's finer decomposition of tasks. 
Using our SDVP on GQA, the performance improvement on InternVL-1.5 is less compared to BLIP. We believe this is primarily because the GQA dataset is already in the fine-tuning stage of the General QA task for InternVL-1.5~\cite{chen2024fargpt4vclosinggap}, which diminishes the performance gains obtained through stepwise distillation on the GQA dataset using our SDVP.

\subsubsection{\textbf{Performance Gain after StepWise Distillation on NLVRv2}}
After stepwise distillation on GQA, we try to distill the knowledge of another Teacher Beit3-NLVRv2 (Beit3 fine-tuned on NLVRv2) on NLVRv2 into the abovementioned visual sub-modules in VisProg and ViperGPT. In Tab.~\ref{tab:SDVP_GQA_NLVRv2}, we take ``SDVP$_G$", the VProg frameworks have been stepwise distilled on GQA using our SDVP, as our baseline; NLVRv2 as the target VR task. It is evident that invoking the teacher model Beit3-NVLRv2 as corresponding visual sub-modules leads to an increase of 8.3\% and 5\% compared with the VisProg baseline, 5.6\% and 4.2\% compared with the ViperGPT baseline on the public test split of the target task NLVRv2. 
However, this improvement comes at the expense of a significant decrease in performance on other tasks, with a maximum decline of 29.6\% on GQA compared with the Visprog baseline. In contrast, our method (SDVP$_{GN}$) not only achieves significant performance improvements on the target VR task (NLVRv2) but also shows minimal declines (anti-forgetting phenomenon) in performance on previous and other tasks. 
This demonstrates that our SDVP can serve as a continual learning strategy for VProg on VR tasks: enhancing performance on target VR tasks and also maintaining a promising performance on un-seen and previous tasks, while preserving the interpretability and flexibility of this framework. 

\subsection{Ablation Study}
\subsubsection{\textbf{Different Dataset Sizes for SDVP}}
 To evaluate the impact of train data volume used in our SDVP, we sampled three different-sized subsets from the GQA train\_balance dataset at equal intervals for our SDVP experiments. Specifically, for VisProg, the sizes of the data subsets sampled are 2,000, 10,000, and 140,000; for ViperGPT, the sizes are 2,000, 10,000, and 80,000. The corresponding experimental results are recorded in Tab.~\ref{tab:ABlation_Data_Size_SDVP_GQA}. Here ``VisProg\_Baseline" means VisProg with pre-trained BLILP invoked as VQA module. ``VisProg\_SDVP$_G$" means ``VisProg\_Baseline" stepwise distilled on GQA using our SDVP (similar for ``ViperGPT\_Baseline" and ``ViperGPT\_SDVP$_G$). As shown in Tab.~\ref{tab:ABlation_Data_Size_SDVP_GQA}, the performance of our SDVP on the GQA dataset increases with the amount of data used, regardless of the VProg framework employed. Moreover, the effect under ViperGPT is more sensitive to increases in data volume. This sensitivity may be attributed to the finer granularity of visual programming in ViperGPT, where the same number of original questions can generate a larger amount of sub-task data for stepwise distillation, and the gap between training and testing split is relatively smaller. 

\begin{table}[h]
    \caption{Results of our SDVP on different scales of distillation data size on GQA compared with baseline. 
    }
    \footnotesize
\centering
    \begin{tabular}{c|c|c}
    \hline
     \multirow{3}{*}{Method}  &\multirow{3}{*}{Data Size} & \textbf{GQA}(Distilled on) \\
     &  & test-dev balanced \\
     \hline
     &  & Accuracy(\%)\\
     \hline
      VisProg\_Baseline & - & 45.1 \\
     \hline
     \multirow{3}{*}{VisProg\_SDVP$_G$}     &2,000& 45.9(+0.8) \\
     &10,000 & 46.5(+1.4) \\
     &140,000 & 47.5(+2.4) \\
     \hline
     ViperGPT\_Baseline & - & 43.1 \\
     \hline
     \multirow{3}{*}{ViperGPT\_SDVP$_G$} &2,000& 46.0(+2.9) \\
     &10,000 & 47.8(+4.7) \\
     &80,000 & 49.2(+6.1) \\
     \hline
    \end{tabular}
    \label{tab:ABlation_Data_Size_SDVP_GQA}
\end{table}

\subsubsection{\textbf{Different Number of Visual Sub-modules Distilled}}
To evaluate the effectiveness of distilling different numbers and types of visual sub-modules using our SDVP, We test ViperGPT with parts of visual sub-modules distilled through our SDVP on GQA. As shown in Tab.~\ref{tab:ABlation_distill_num_SDVP_GQA}, distilling different types and numbers of visual sub-modules can yield varying effects. For the stepwise distillation of a single visual sub-module, distilling the ``simple\_query" module results in the most significant improvement, increasing accuracy by 6.2\%. This may be due to the ``simple\_query" being called more frequently (63.5\% of the time) in the visual programs on GQA as compared to the other two sub-modules used by ViperGPT. Distilling all three types of visual sub-modules achieves the highest improvement (+6.5\%). However, we also observed that sometimes distilling two sub-modules results in lower performance than distilling just one, which might be due to the use of pseudo-labels for stepwise distillation and the inherent code error in ViperGPT. Further detailed analysis of these reasons will be left for future work.

\begin{table}[h]
    \caption{Results of stepwise distilling different numbers of visual sub-modules of ViperGPT on GQA with our SDVP. sp, bt, and vp respectively denote the ``simple\_query", ``best\_text\_match" and ``verify\_property" sub-modules of ViperGPT. Checking the box means that the corresponding visual sub-modules are stepwise distilled using our SDVP. }
    \footnotesize
\centering
    \begin{tabular}{c|ccc|c}
    \hline
     \multirow{3}{*}{Method}  &\multirow{3}{*}{sp} &\multirow{3}{*}{bt} &\multirow{3}{*}{vp}& \textbf{GQA}(Distilled on) \\
     &  & & & test-dev balanced \\
     \hline
     &  & & & Accuracy(\%)\\
     \hline
      ViperGPT\_Baseline &  &  & & 43.1 \\
     \hline
     \multirow{7}{*}{ViperGPT\_SDVP$_G$}     &\Checkmark   &  &  & 49.3(+6.2) \\
     &   & \Checkmark &  & 43.8(+0.7) \\
     &   &  & \Checkmark & 45.8(+2.7)  \\
     & \Checkmark  & \Checkmark &  & 48.4(+5.3)  \\
     & \Checkmark  &  & \Checkmark & 49.3(+6.2)  \\
     &   & \Checkmark & \Checkmark & 45.0(+1.9)  \\
     & \Checkmark  & \Checkmark & \Checkmark & 49.6(+6.5)  \\
     
     \hline
    \end{tabular}
    \label{tab:ABlation_distill_num_SDVP_GQA}
\end{table}

\subsubsection{\textbf{Generalization Across VProg Frameworks}}
Different VProg frameworks vary in code style and granularity. ViperGPT has a finer granularity than VisProg, including control branches such as ``if" and ``for". Here we explore whether the visual model selected to be distilled using our SDVP has the cross-framework generalization ability, which means that when another VProg framework invoking this visual model can also enhance the performance of the target VR task. To evaluate it, we use our SDVP to fine-tune two different VProg frameworks (VisProg and ViperGPT) on GQA with the visual model (here we use pre-trained BLP) for the ``VQA" module in VisProg and ``verify\_property" ``best\_test\_match" and ``simple\_query" modules in ViperGPT been stepwise distilled under each VProg framework respectively. Then we make VisProg invoke the visual model distilled under ViperGPT as its ``VQA" module (VisProg$_{ViperGPT}$) and make ViperGPT invoke the visual model distilled under VisProg as its ``verify\_property" ``best\_text\_match" ``simple\_query" modules (ViperGPT$_{VisProg}$) respectively and evaluate the accuracy of these two VProg frameworks on the GQA test-dev balanced split and the NLVRv2 public test split. The performances are recorded in Tab.~\ref{tab:ABlation_cross_framework_SDVP_GQA_NLVRv2}. From Tab.~\ref{tab:ABlation_cross_framework_SDVP_GQA_NLVRv2} we can see that even using visual models stepwise distilled under another VProg framework can also enhance the performance on the target VR task, and effectively prevent forgetting or even endow a transfer learning gain on other tasks such as NLVRv2. Those experimental results demonstrate the cross-framework generalization ability of the visual models distilled using our SDVP. 


\begin{table}[h]
    \caption{Generalization of our SDVP across VProg frameworks. VisProg$_{ViperGPT}$ represents VisProg with relevant visual sub-modules invoking models distilled under ViperGPT with our SDVP. ViperGPT$_{VisProg}$ represents ViperGPT with relevant visual sub-modules invoking models distilled under VisProg with our SDVP.}
    \footnotesize
\centering
    \begin{tabular}{c|c|c}
    \hline
     \multirow{3}{*}{Method}   & \textbf{GQA}(Distilled on) & \textbf{NLVRv2}\\
     &test-dev balanced  & public test\\
     \hline
     & Accuracy(\%) & Accuracy(\%)\\
     \hline
      VisProg\_Baseline & 45.1 &65.7\\
      VisProg$_{ViperGPT}$ & 47.1(+2.0) & 65.7(-0.0) \\
     \hline
    
     ViperGPT\_Baseline & 43.1  & 60.5 \\
     ViperGPT$_{VisProg}$ & 49.6(+6.5) & 65.8(+5.3) \\
     \hline
    \end{tabular}
    \label{tab:ABlation_cross_framework_SDVP_GQA_NLVRv2}
\end{table}

\subsection{Case Study}

\begin{figure*}[t]
\centering
\includegraphics[width=\textwidth]{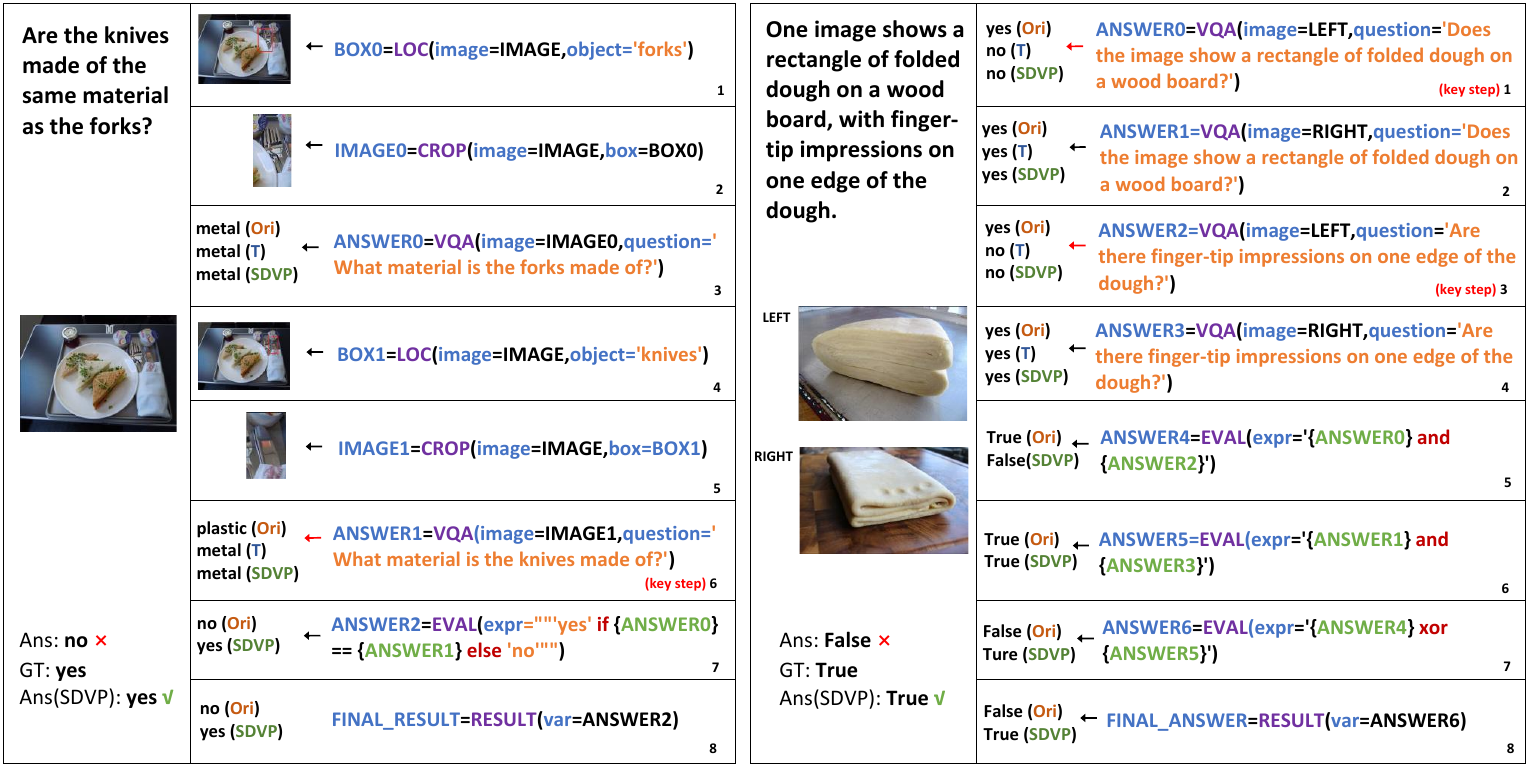}
\vspace{-22pt}
\caption{Qualitative examples showing the program execution flow of VisProg original and after learning with our SDVP. For each example, we provide detailed process outputs during execution and the final answers, in which Ans means the final answer, GT means the ground truth, Ori means the original VisProg, T means the teacher, SDVP means VisProg after learning with our SDVP.
}
\vspace{-10pt}
\label{fig_visprog_case}
\end{figure*}


\begin{figure*}[htb]
	
	\begin{minipage}{0.5\linewidth}
		\vspace{3pt}
		\centerline{\includegraphics[width=\textwidth]{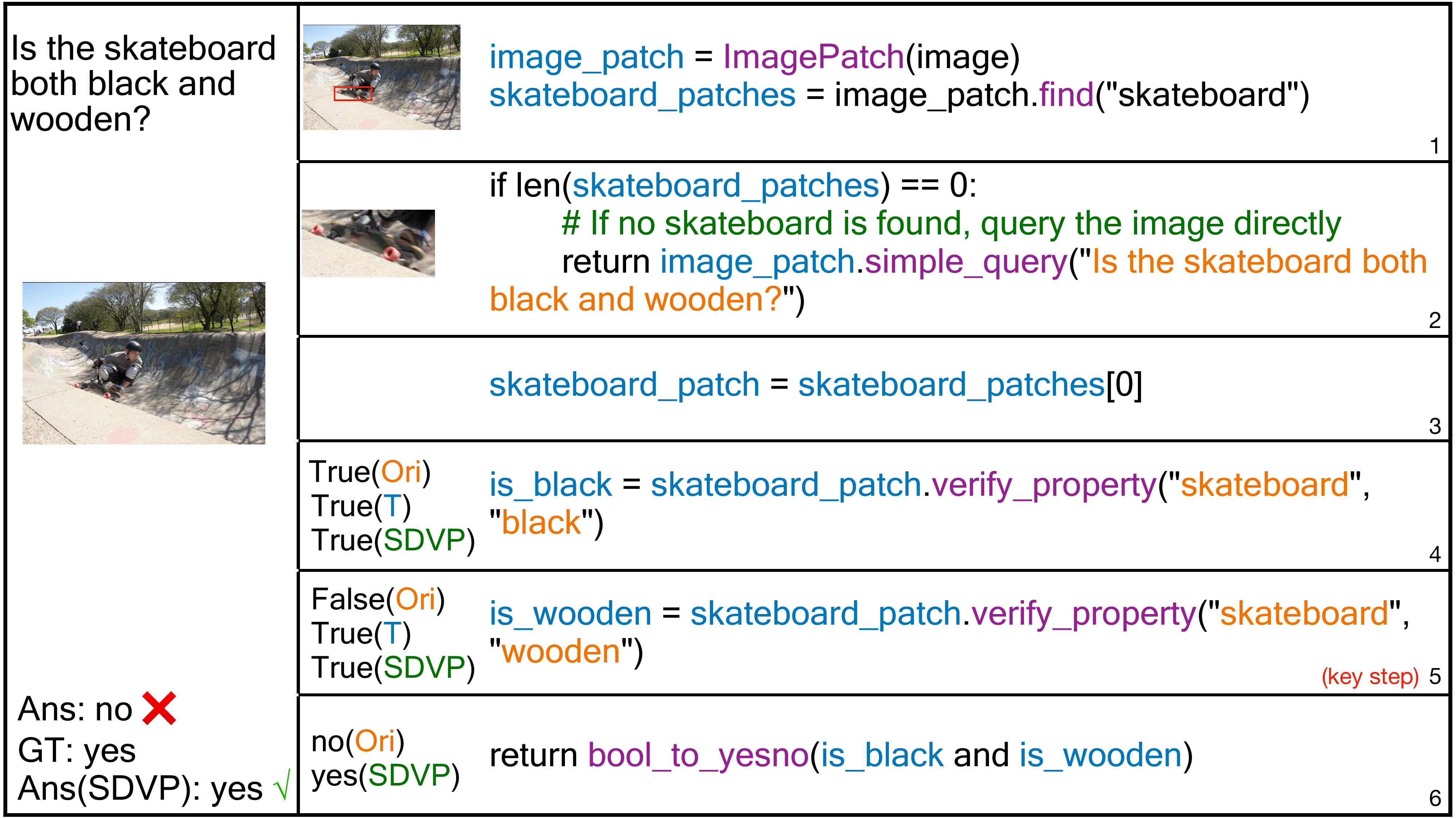}}
		
	\end{minipage}
	\begin{minipage}{0.492\linewidth}
		\vspace{3pt}
		\centerline{\includegraphics[width=\textwidth]{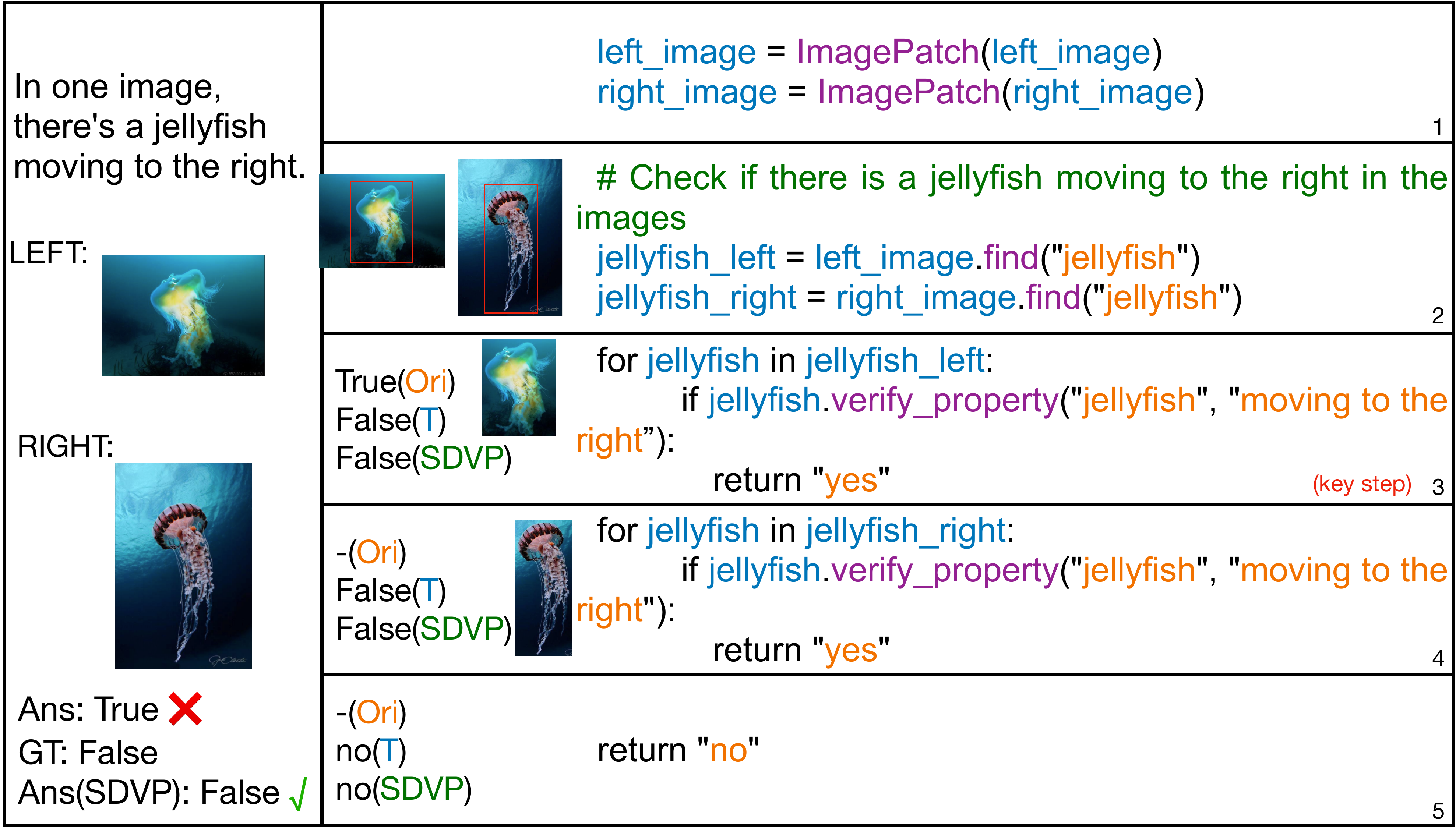}}
		
	\end{minipage}
\caption{Qualitative examples showing the program execution flow of ViperGPT original and after learning with our SDVP. For each example, we provide detailed process outputs during execution and the final answers, in which Ans means the final answer, GT means the ground truth, Ori means the original ViperGPT, T means the teacher, SDVP means ViperGPT after learning with our SDVP.
}
\label{fig_viper_case}
\end{figure*}

We provide qualitative examples in Fig.~\ref{fig_visprog_case} to show how SDVP works for VisProg. In the first example, facing the image and the question ``Are the knives made of the same material as the forks?", VisProg first uses the visual sub-module ``LOC" to find the forks, and crop them for the further visual sub-question. Then the ``VQA"  module is invoked to answer what material the forks are made of. The same operations are executed for the knives. However, the ``VQA" module gives a wrong answer ``plastic" as the material of the knives, which causes a non-correct final answer ``no". We use SDVP to stepwise distill the Teacher's answers to those visual sub-questions to the ``VQA" module. After distillation, the ``VQA" module answers ``metal" correctly in the key step, and we get the answer to the original question ``yes", which is the same as the ground truth. 
In the second example, VisProg invokes the visual sub-module ``VQA" to answer if a rectangle of folded dough on a wood board, with fingertip impressions on one edge of the dough in only one image. Original VisProg gets the wrong final answer ``False". After employing our SDVP, the answers to the first and third steps have been corrected so that the final answer has been corrected from ``False" to ``True". Similarly, in Fig.~\ref{fig_viper_case}, we show how our SDVP helps to enhance the performance of ViperGPT through specific examples.
\section{Discussion}

\subsection{Error Analysis}
\label{subsec: error_analysis}
\begin{figure}[]
\centering
    \includegraphics[width=1\columnwidth]{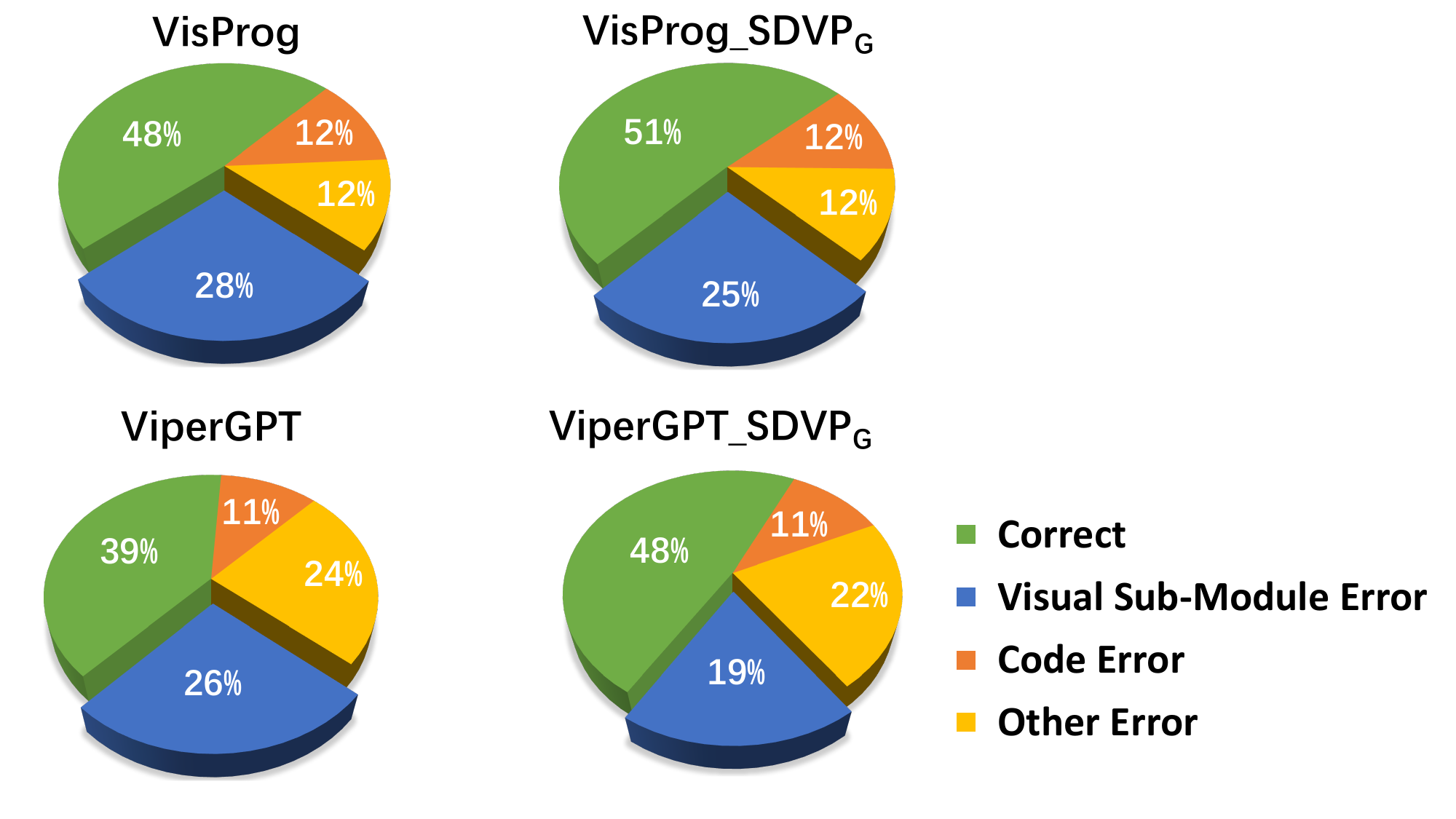}
\vspace{-20pt}
\caption{Sources of errors in VisProg and ViperGPT before and after stepwise distilling on GQA with our SDVP. Here, `Other Error' includes synonyms, acceptable answers, and dirty data.  
}
\vspace{-10pt}
\label{fig5}
\end{figure}

To validate whether our method is effective in improving the performance of visual sub-modules in VProg, we manually inspect visual programs of 100 randomly selected examples in the GQA test-dev balanced split to analyze the source of errors. As is depicted in Fig.~\ref{fig5}, 
the primary source of errors in VisProg stems from faults in the visual sub-module. The error rate of the visual sub-module in SDVP decreased by 3\%, leading to an overall accuracy improvement of 3\%. Similarly, we also observed the same trend in ViperGPT. After stepwise distillation, the error rate of the visual sub-modules decreased by 7\%. Besides, 2\% of errors caused by synonyms can also be corrected after distillation. The decrease in error rates of visual sub-modules error and other errors leads to an overall accuracy improvement of 9\%. 
This demonstrates that our SDVP enhances the performance of VProg on target tasks by enhancing the capabilities of visual sub-modules invoked.

\subsection{Why our SDVP can alleviate ``Catastrophic Forgetting"}
\label{subsec: alleviate forgetting}
The experimental results in Tab.~\ref{tab:SDVP_GQA} and~\ref{tab:SDVP_GQA_NLVRv2} reveal that after applying our SDVP for stepwise distillation on the visual sub-modules of VProg in the target VR task, VProg exhibits very little performance degradation on other tasks, indicating a strong resistance to forgetting. In this section, we conduct detailed analyses of the reasons behind this phenomenon. We guess that the primary reason is that VProg decouples complex composite VR tasks, thereby reducing the gap in input distributions between different tasks (i.e., the input gap for the same visual sub-tasks performed by the same visual modules across different tasks becomes smaller). This, in turn, enables the learning strategy based on it (our SDVP) and the inference strategy (executing the visual program of VProg) to exhibit stronger resistance to forgetting compared to the original learning and inference strategies based on pure end-to-end neural network models. Next, we will attempt to analyze the abovementioned guess from intuitive qualitative and quantitative experimental perspectives.
\begin{figure}[t]
\centering
\includegraphics[width=\columnwidth]{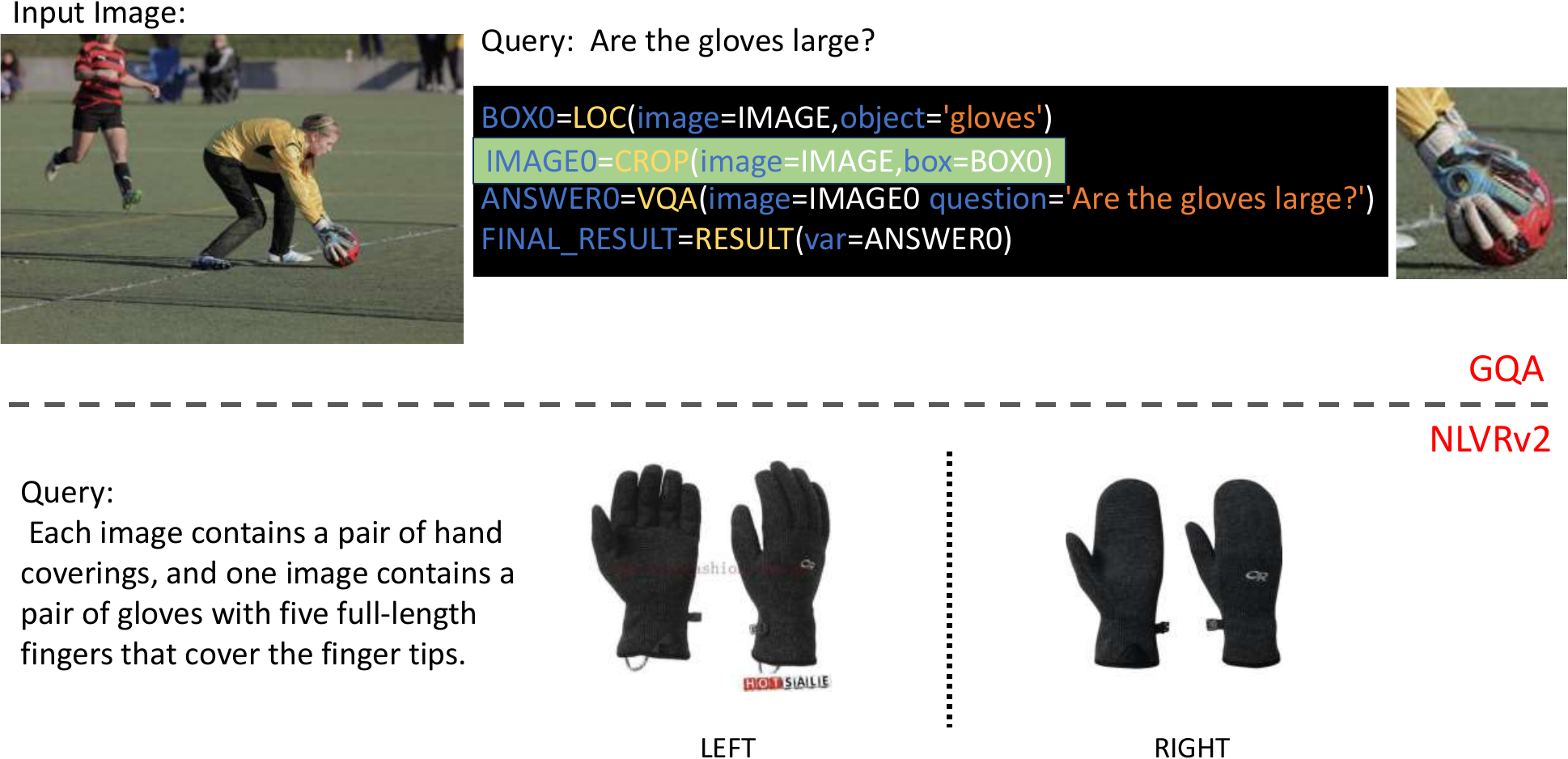}
\caption{Examples of original input images in the GQA and NLVRv2 dataset and cropped images in Visprog. In this example, the cropped image shows a pair of gloves holding a ball, which appears more similar to the images in NLVRv2 compared to the original image.}
\vspace{-15pt}
\label{fig:domain_gap}
\end{figure}

Here, we illustrate through intuitive examples that, compared to fine-tuning the original task data, the input domain gap in different tasks for the VQA Module in the stepwise distillation of our SDVP may be smaller. This is manifested as the inputs of previous and new tasks are more similar. From Fig.~\ref{fig:domain_gap} we can see, compared with the original input image in this case, after the ``crop" manipulation, the input sub-image of the VQA Module in Visprog (one kind of VProg framework) looks more similar to the below input image of the case in the NLVRv2 task. 

The same thing happened for the question (or text) inputs. We speculate that the similarity between the sub-questions on GQA and NLVRv2, which are decoupled through VProg, would be higher than the similarity between the original composite long-form questions on these two tasks. For instance, there exists an original question in GQA: ``Are the trees brown and abundant?" and another original statement in NLVRv2: ``There are two trees in the image on the right." When these two questions are decoupled by VisProg (one kind of VProg framework), the former question becomes two separate ones: \{``Are the trees brown?", ``How many trees are there?"\}. The latter question is decoupled into a single sub-question: ``How many trees are in the image?", which is very similar to the second sub-question decomposed from the former GQA question. Intuitively, the sub-questions after decoupling are more similar between these two examples from different tasks.

To further clarify our analyses, we conducted an external experiment. This experiment started with the BLIP\cite{blip} model fine-tuned on VQA\cite{antol2015vqa}, we measured its performance on the original VQA and GQA dataset inputs. (We do not use the NLVRv2 dataset here because its end-to-end neural network takes two images, which results in an inconsistent interface, making it inconvenient to directly measure the forgetting on end-to-end networks when learning sequentially on GQA and NLVRv2.) We then separately adopted fine-tuning on the original dataset questions and the stepwise distillation technique in our SDVP for learning on GQA, and then respectively measured the performance of the model when using the VQA and GQA original datasets as inputs, as well as the performance of the model as VQA module in an interpretable VisProg on the VQA (randomly drew ten thousand samples from the val\_set) and GQA (the entire test-dev balanced split) datasets. The results are recorded in Tab.~\ref{tab:why}.

\begin{table}
    \caption{A comparison of the degree of forgetting between direct fine-tuning and stepwise distillation. ``ori\_input" means using original task data as input, ``under\_VisProg" means getting the accuracy using VisProg which invokes the learned BLIP model as the VQA Module. ``\textbf{BLIP\_ori}" means the BLIP model just fine-tuned on VQA, and have not learned on GQA yet. ``\textbf{BLIP\_FT$_{G\_ori}$}" means the BLIP model fine-tuned on GQA original data after fine-tuned on VQA. ``\textbf{BLIP\_SD$_G$}" means the BLIP model is stepwise distilled on GQA after fine-tuned on VQA.}
    \begin{adjustbox}{width=\columnwidth}
    \centering
    \renewcommand{\arraystretch}{1.4}
    \begin{tabular}{c|cccc}
    \hline
    \multirow{2}{*}{Method}    &\multicolumn{2}{c}{\textbf{VQA}(previous task)} & \multicolumn{2}{c}{\textbf{GQA}(target task)}  \\
     & ori\_input & under\_VisProg & ori\_input & under\_VisProg \\
    \hline
    \textbf{BLIP\_ori} & 85.4 & 72.6 & 49.2 & 45.1 \\
    \hline
    \textbf{BLIP\_FT$_{G\_ori}$} & 70.7(-14.7) & - & 50.9(+1.7) & - \\
    \hline
    \textbf{BLIP\_SD$_G$} & - & 65.1(-7.5) & - & 47.5(+2.4) \\
    \hline
    \end{tabular}
    \end{adjustbox}
    \label{tab:why}
\end{table}

From the table, we can see that after directly fine-tuning the original GQA data, the accuracy of BLIP on original GQA inputs increased by 1.7\% (the limited increase is due to the smaller scale of GQA training data used for aligning with the stepwise distillation experiment, as the stepwise distillation experiment requires code generated by LLM, which is costly, so we limited the scale of the training set). Meanwhile, the accuracy of the original data of the previous task VQA dropped by 14.7 percentage points, showing a significant decline.

Considering the results under the interpretable framework—VisProg, using the stepwise distillation technique in our SDVP can help improve the performance of VisProg in GQA from 45.1\% to 47.5\% (an increase of 2.4 percentage points). At the same time, the accuracy on the previously learned VQA task dropped from 72.6\% to 65.1\% (only a decline of 7.5 percentage points), demonstrating stronger resistance to forgetting compared to the control group.


Our preliminary observation led us to speculate that the inputs of sub-steps after decomposing by VProg across different tasks are more similar than the original inputs, which might lead to its anti-forgetting feature. However, the aforementioned statement is only our speculation and requires further investigation, including more rigorous formal proofs or comprehensive sample analyses based on systematic statistical methods. We hope our discussion here provides some relevant insights. 

\subsection{Limitation}
Though SDVP shows anti-forgetting capability, its ability to improve performance on target tasks is limited by the ability of teacher models since there are only pseudo labels generated by teacher models involved in our method. Moreover, as SDVP is a learning strategy for VProg that utilizes visual programs generated by LLMs, SDVP also inherits the risks of LLMs, such as outputting programs with syntax or semantic errors.

\section{Conclusion}

We propose SDVP, a continual learning strategy for non-differentiable VProg across various VR tasks. Our SDVP stepwise distills the process knowledge from an existing task-specific model into visual sub-modules of VProg, enabling VProg to improve performance across various specific VR tasks while maintaining performance on unseen and previous tasks. Experiments across various VR tasks have demonstrated the effectiveness of our SDVP. Our planned future work is divided into two directions: one is to optimize the visual sub-modules and visual programs within VProg collaboratively, and the other direction is to explore how to reduce reliance on the teacher model and directly use the final supervision to enhance the VProg framework.

\bibliographystyle{IEEEtran}
\bibliography{SDVP}












\section{Biography Section}

\vspace{11pt}

\begin{IEEEbiography}[{\includegraphics[width=1in,height=1.25in,clip,keepaspectratio]{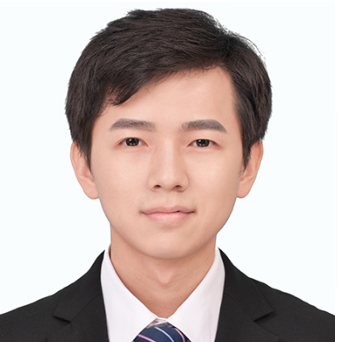}}]{Wentao Wan}
is currently a Ph.D candidate at the School of Computer Science and Engineering, Sun Yat-sen University, since 2019. He received the M.S. degree in Computer Science and Technology from Huazhong University of Science and Technology
in 2018, and the B.S. degree from Central South University in 2015. His main research interests include Visual Reasoning, Multi-modal Learning, General Reasoning, and Continual Learning.
\end{IEEEbiography}

\vspace{11pt}

\begin{IEEEbiography}[{\includegraphics[width=1in,height=1.25in,clip,keepaspectratio]{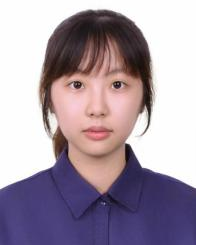}}]{Nan Kang} received the Bachelor’s degree in Software Engineering from Central South University, Changsha, China, in 2024. She is currently a graduate student in progress at the School of Computer Science and Engineering, Sun Yat-sen University, Guangzhou, China. Her research interests include multi-modal reasoning, visual programming, and chain-of-thought.
\end{IEEEbiography}

\vspace{11pt}

\begin{IEEEbiography}[{\includegraphics[width=1in,height=1.25in,clip,keepaspectratio]{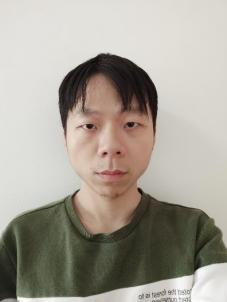}}]{Zhuojie Yang} received his Bachelor's degree in Computer Science and Technology from Sun Yat-sen University in 2023. He is currently a second-year master's student in Computer Technology at the same university and a member of the HCP Lab. His research interests primarily focus on large language models and multimodal systems, exploring advanced methods in artificial intelligence to enhance model performance and application in various fields.
\end{IEEEbiography}

\vspace{11pt}

\begin{IEEEbiography}[{\includegraphics[width=1in,height=1.25in,clip,keepaspectratio]{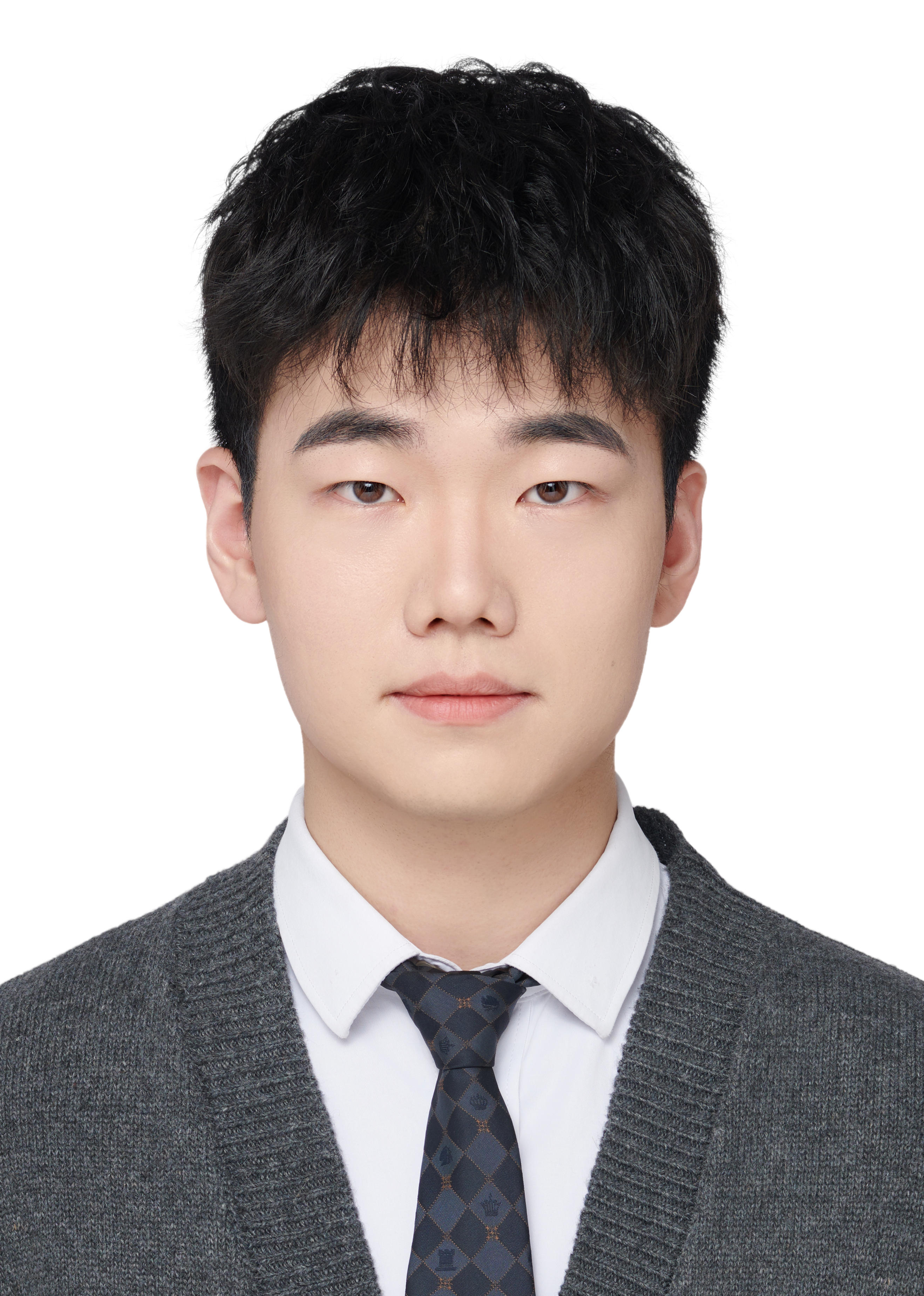}}]{Zeqing Wang} received the B.S. degree of software engineering from the College of Software, Jilin University, Changchun, China in 2023. He is currently a M.S. student at the School of Computer Science and Engineering, Sun Yat-sen University. His Main interests include multi-modal learning, multi-agent systems, and vision language models. He has been serving as a reviewer for numerous academic journals and conferences, such as TMM, ACM MM.
\end{IEEEbiography}

\vspace{11pt}

\begin{IEEEbiography}[{\includegraphics[width=1in,height=1.25in,clip,keepaspectratio]{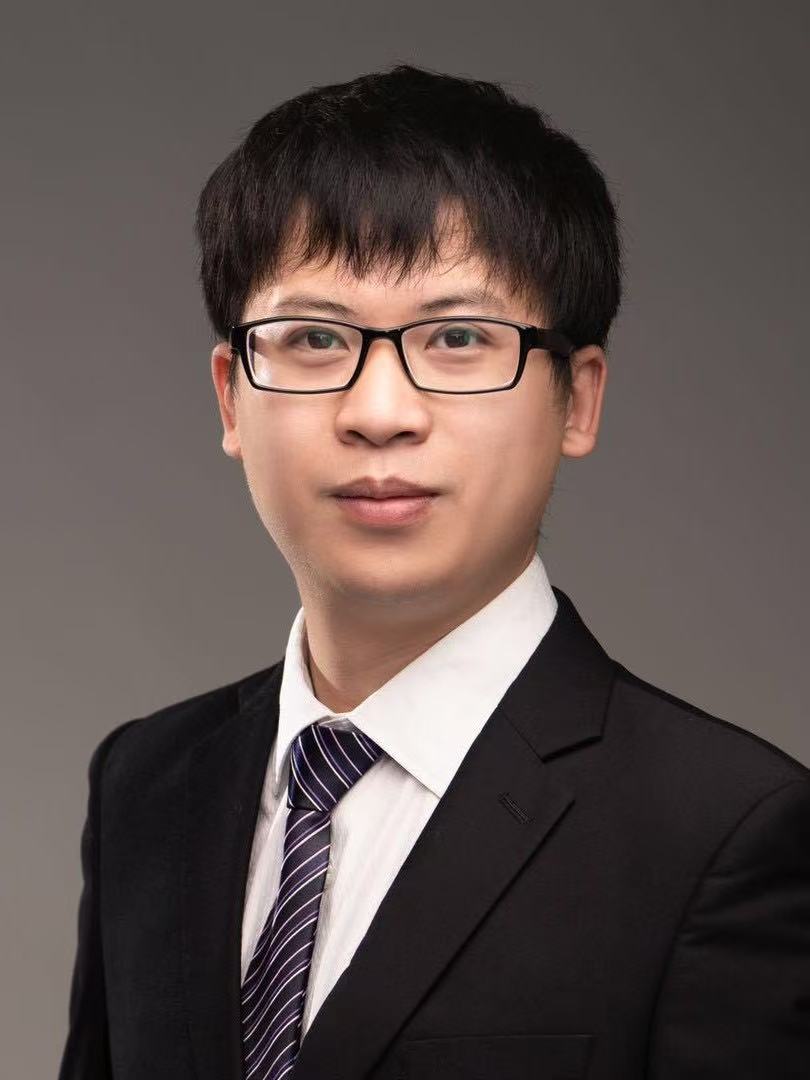}}]{Keze Wang} is nationally recognized as the Distinguished Young Scholars, currently serving as an Associate Professor at the School of Computer Science, Sun Yat-sen University, and a doctoral supervisor. He holds two Ph.D. degrees from Sun Yat-sen University (2017) and the Hong Kong Polytechnic University (2019). In 2018, he worked as a postdoctoral researcher at the University of California, Los Angeles, and returned to Sun Yat-sen University in 2021 as part of the "Hundred Talents Program." Dr. Wang has focused on reducing deep learning's dependence on training samples and mining valuable information from massive unlabeled data.
\end{IEEEbiography}

\vspace{11pt}

\begin{IEEEbiography}[{\includegraphics[width=1in,height=1.25in,clip,keepaspectratio]{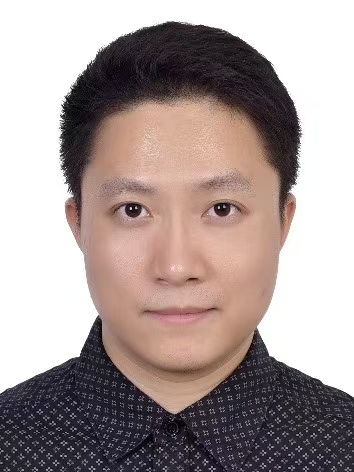}}]{Liang Lin}(IEEE Fellow) is a full professor of computer science with Sun Yat-sen University. He served as the executive director and distinguished scientist of SenseTime Group from 2016 to 2018, leading the R\&D teams for cutting-edge technology transferring. He has authored or co-authored more than 200 papers in leading academic journals and conferences, and his papers have been cited by more than 26\,000 times. He is an associate editor of \textit{IEEE Trans. Neural Networks and Learning Systems} and \textit{IEEE Trans. Multimedia}, and served as area chairs for numerous conferences, such as CVPR, ICCV, SIGKDD, and AAAI. He is the recipient of numerous awards and honors including Wu Wen-Jun Artificial Intelligence Award, the First Prize of China Society of Image and Graphics, ICCV Best Paper Nomination, in 2019, Annual Best Paper Award by Pattern Recognition (Elsevier), in 2018, Best Paper Dimond Award in IEEE ICME 2017, Google Faculty Award, in 2012. His supervised PhD students received ACM China Doctoral Dissertation Award, CCF Best Doctoral Dissertation and CAAI Best Doctoral Dissertation. He is a fellow of IET/IAPR.
\end{IEEEbiography}

\vfill

\end{document}